# Oblivious Bounds on the Probability of Boolean Functions

WOLFGANG GATTERBAUER, Carnegie Mellon University
DAN SUCIU, University of Washington

This paper develops upper and lower bounds for the probability of Boolean functions by treating multiple occurrences of variables as independent and assigning them new individual probabilities. We call this approach *dissociation* and give an exact characterization of *optimal oblivious bounds*, i.e. when the new probabilities are chosen independent of the probabilities of all other variables. Our motivation comes from the *weighted model counting* problem (or, equivalently, the problem of computing the probability of a Boolean function), which is #P-hard in general. By performing several dissociations, one can transform a Boolean formula whose probability is difficult to compute, into one whose probability is easy to compute, and which is guaranteed to provide an upper or lower bound on the probability of the original formula by choosing appropriate probabilities for the dissociated variables. Our new bounds shed light on the connection between previous relaxation-based and model-based approximations and unify them as concrete choices in a larger design space. We also show how our theory allows a standard relational database management system (DBMS) to both upper and lower bound hard probabilistic queries in guaranteed polynomial time.

## 1. INTRODUCTION

Query evaluation on probabilistic databases is based on weighted model counting for positive Boolean expressions. Since model counting is #P-hard in general, today's probabilistic database systems evaluate queries using one of the following three approaches: (1) incomplete approaches identify tractable cases (e.g., read-once formulas) either at the query-level [Dalvi and Suciu 2007; Dalvi et al. 2010] or the data-level [Olteanu and Huang 2008; Sen et al. 2010]; (2) exact approaches apply exact probabilistic inference, such as repeated application of Shannon expansion [Olteanu et al. 2009] or tree-width based decompositions [Jha et al. 2010]; and (3) approximate approaches either apply general purpose sampling methods [Jampani et al. 2008; Kennedy and Koch 2010; Re et al. 2007] or approximate the number of models of the Boolean lineage expression [Olteanu et al. 2010; Fink and Olteanu 2011].

This paper provides a new *algebraic framework* for approximating the probability of positive Boolean expressions. While our method was motivated by query evaluation on probabilistic databases, it is more general and applies to all problems that rely on weighted model counting, e.g., general probabilistic inference in graphical mod-els [Chavira and Darwiche 2008].[1] An important aspect of our method is that it is *not model-based* in the traditional sense. Instead, it enlarges the original variable space by treating multiple occurrences of variables as independent and assigning them new individual probabilities. We call this approach *dissociation*[2] and explain where existing relaxation-based and model-based approximations fit into this larger space of approximations. We characterize probability assignments that lead to guaranteed upper or lower bounds on the original expression and identify the best possible *oblivious* bounds, i.e. after looking at only a limited scope of the expression. We prove that for every model-based bound there is always a dissociation-based bound that is *as good or better*. And we illustrate how a standard relational DBMS can both upper and lower bound hard probabilistic conjunctive queries without self-joins with appropriate SQL queries that use dissociation in a query-centric way.

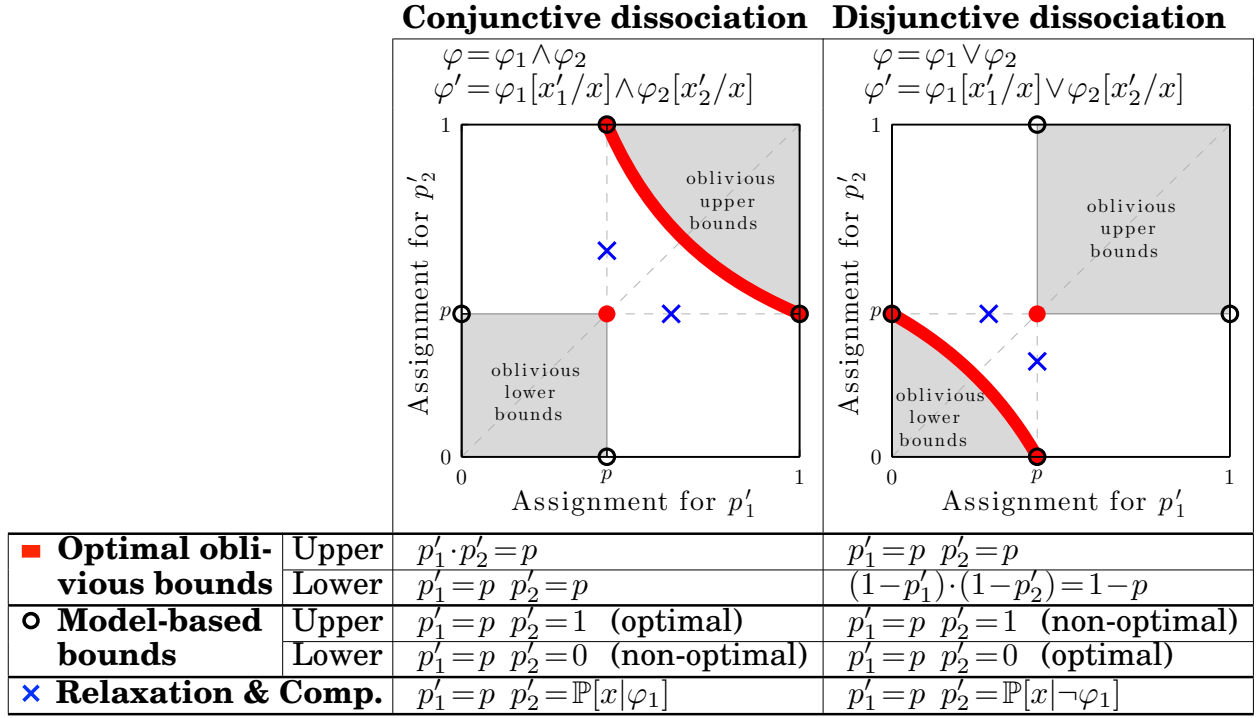

**Fig. 1.** Dissociation as framework that allows to determine optimal oblivious upper and lower bounds for the probabilities $\mathbf{p}' = \langle p'_1, p'_2 \rangle$ of dissociated variables. Oblivious here means that we assign new values after looking at only a limited scope of the expression. *Model-based* upper conjunctive and lower disjunctive bounds are obliviously optimal (they fall on the red line of optimal assignments), whereas lower conjunctive and upper disjunctive are not. *Relaxation & Compensation* is a form of dissociation which is not oblivious ($p_2$ is calculated with knowledge of $\varphi_1$) and does not, in general, guarantee to give an upper or lower bound.

We briefly discuss our results: We want to compute the probability $\mathbb{P}[\varphi]$ of a Boolean expression $\varphi$ when each of its Boolean variables $x_i$ is set independently to true with some given probability $p_i = \mathbb{P}[x_i]$. Computing $\mathbb{P}[\varphi]$ is known to be #P-hard in general [Valiant 1979] and remains hard to even approximate [Roth 1996]. Our approach is to approximate $\mathbb{P}[\varphi]$ with $\mathbb{P}[\varphi']$ that is easier to compute. The new formula $\varphi'$ is derived from $\varphi$ through a sequence of *dissociation steps*, where each step replaces $d$ distinct occurrences of some variable $x$ in $\varphi$ with $d$ fresh variables $x'_1, x'_2, \ldots x'_d$. Thus, after applying dissociation repeatedly, we transform $\varphi$ into another expression $\varphi'$ and approximate $\mathbb{P}[\varphi]$ with $\mathbb{P}[\varphi']$. The question that we address in this paper is: how should we set the probabilities of the dissociated variables $x_i'$ in order to ensure that $\mathbb{P}[\varphi']$ is a good approximation of $\mathbb{P}[\varphi]$? In particular, we seek conditions under which $\varphi'$ is guaranteed to be either an upper bound $\mathbb{P}[\varphi'] \geq \mathbb{P}[\varphi]$ or a lower bound $\mathbb{P}[\varphi'] \leq \mathbb{P}[\varphi]$.

---

[1]Note that *weighted model counting* is essentially the same problem as computing the probability $\mathbb{P}[\varphi]$ of a Boolean expression $\varphi$. Each truth assignment of the Boolean variables corresponds to one model whose weight is the probability of this truth assignment. Weighted model counting then asks for the sum of the weights of all satisfying assignments.

[2]*Dissociation* is the breaking of an existing association between related, but not necessarily identical items.

Our main result can be summarized as follows: Suppose that $x$ occurs positively in $\varphi$. Dissociate $x$ into two variables $x_1'$ and $x_2'$ such that the dissociated formula is $\varphi' = \varphi_1' \wedge \varphi_2'$, where $x_1'$ occurs only in $\varphi_1'$ and $x_2'$ occurs only in $\varphi_2'$; in other words, $\varphi \equiv \varphi_1'[x_1'/x] \wedge \varphi_2'[x_2'/x]$ (we will later define this as "conjunctive dissociation"). Let $p = \mathbb{P}[x]$, $p_1' = \mathbb{P}[x_1']$, $p_2' = \mathbb{P}[x_2']$ be their probabilities. Then (1) $\mathbb{P}[\varphi'] \geq \mathbb{P}[\varphi]$ iff $p_1' \cdot p_2' \geq p$, and (2) $\mathbb{P}[\varphi'] \leq \mathbb{P}[\varphi]$ iff $p_1' \leq p$ and $p_2' \leq p$. In particular, the best upper bounds are obtained by choosing any $p_1', p_2'$ that satisfy $p_1' \cdot p_2' = p$, and the best lower bound is obtained by setting $p_1' = p_2' = p$. The "only if" direction holds assuming $\varphi'$ satisfies certain mild conditions (e.g., it should not be redundant), and under the assumption that $p_1', p_2'$ are chosen *obliviously*, i.e. they are functions only of $p = \mathbb{P}[x]$ and independent of the probabilities of all other variables. This restriction to oblivious probabilities guarantees the computation of the probabilities $p_1', p_2'$ to be very simple.[3] Our result extends immediately to the case when the variable $x$ is dissociated into several variables $x_1', x_2', \ldots, x_d'$, and also extends (with appropriate changes) to the case when the expressions containing the dissociated variables are separated by $\vee$ rather than $\wedge$ (Fig. 1).

*Example* 1.1 (*2CNF Dissociation*). For a simple illustration of our main result, consider a Positive-Partite-2CNF expression with $|E|$ clauses

$$\varphi = \bigwedge_{(i,j)\in E} (x_i \vee y_j) \tag{1}$$

for which calculating its probability is already #P-hard [Provan and Ball 1983]. If we dissociate all occurrences of all $m$ variables $x_i$, then the expression becomes:

$$\varphi' = \bigwedge_{(i,j)\in E} (x_{i,j}' \vee y_j) \tag{2}$$

which is equivalent to $\bigwedge_j \left(y_j \vee \bigwedge_{i,j} x_{i,j}'\right)$. This is a read-once expression whose probability can always be computed in PTIME [Gurvich 1977]. Our main result implies the following: Let $p_i = \mathbb{P}[x_i]$, $i \in [m]$ be the probabilities of the original variables and denote $p_{i,j}' = \mathbb{P}[x_{i,j}']$ the probabilities of the fresh variables. Then (1) if $\forall i \in [m] : p_{i,j_1}' \cdot p_{i,j_2}' \cdots p_{i,j_{d_i}}' = p_i$, then $\varphi'$ is an upper bound ($\mathbb{P}[\varphi'] \geq \mathbb{P}[\varphi]$); (2) if $\forall i \in [m] : p_{i,j_1}' = p_{i,j_2}' = \ldots = p_{i,j_{d_i}}' = p_i$, then $\varphi'$ is a lower bound ($\mathbb{P}[\varphi'] \leq \mathbb{P}[\varphi]$). Furthermore, these are the best possible *oblivious* bounds, i.e. where $p_{i,j}'$ depends only on $p_i = \mathbb{P}[x_i]$ and is chosen independently of other variables in $\varphi$. ∎

We now explain how dissociation generalizes two other approximation methods in the literature (Fig. 1 gives a high-level summary and Sect. 5 the formal details).

**Relaxation & Compensation.** This is a framework by Choi and Darwiche [2009; 2010] for approximate probabilistic inference in graphical models. The approach performs exact inference in an approximate model that is obtained by relaxing equiv-alence constraints in the original model, i.e. by removing edges. The framework allows one to improve the resulting approximations by compensating for the relaxed constraints. In the particular case of a conjunctive Boolean formula $\varphi = \varphi_1 \wedge \varphi_2$, *relaxation* refers to substituting any variable $x$ that occurs in both $\varphi_1$ and $\varphi_2$ with two fresh variables $x'_1$ in $\varphi_1$ and $x'_2$ in $\varphi_2$. *Compensation* refers to then setting their probabilities $p'_1 = \mathbb{P}[x'_1]$ and $p'_2 = \mathbb{P}[x'_2]$ to $p'_1 = p$ and $p'_2 = \mathbb{P}[x|\varphi_1]$. This new probability assignment is justified by the fact that, if $x$ is the only variable shared by $\varphi_1$ and $\varphi_2$, then compensation ensures that $\mathbb{P}[\varphi'] = \mathbb{P}[\varphi]$ (we will show this claim in Prop. 5.1). In general, however, $\varphi_1, \varphi_2$ have more than one variable in common, and in this case we have $\mathbb{P}[\varphi'] \neq \mathbb{P}[\varphi]$ for the same compensation. Thus in general, compensation is applied as a heuristics. Furthermore, it is then not known whether compensation provides an upper or lower bound.

---

[3] Our usage of the term *oblivious* is inspired by the notion of oblivious routing algorithms [Valiant 1982] which use only local information and therefore can be implemented very efficiently. Similarly, our *oblivious* framework forces $p_1', p_2'$ to be computed only as a function of $p$, without access to the rest of $\varphi$. One can always find values $p_1', p_2'$ for which $\mathbb{P}[\varphi] = \mathbb{P}[\varphi']$. However, to find those value in general, one has to first compute $q = \mathbb{P}[\varphi]$, then find appropriate values $p_1', p_2'$ for which the equality $\mathbb{P}[\varphi'] = q$ holds. This is not practical, since our goal is to compute $q$ in the first place.

Indeed, let $p'_1 = p$, $p'_2 = \mathbb{P}[x|\varphi_1]$ be the probabilities set by the compensation method. Recall that our condition for $\mathbb{P}[\varphi']$ to be an upper bound is $p'_1 \cdot p'_2 \geq p$, but we have $p'_1 \cdot p'_2 = p \cdot \mathbb{P}[x|\varphi_1] \leq p$. Thus, the compensation method does not satisfy our oblivious upper bound condition. Similarly, because of $p'_1 = p$ and $p'_2 \geq p$, these values fail to satisfy our oblivious lower bound condition. Thus, relaxation is neither a guaranteed upper bound, nor a guaranteed lower bound. In fact, relaxation is not oblivious at all (since $p'_2$ is computed from the probabilities of all variables, not just $\mathbb{P}[x]$). This enables it to be an exact approximation in the special case of a single shared variable, but fails to guarantee any bounds, in general.

**Model-based approximations.** Another technique for approximation described by Fink and Olteanu [2011] is to replace $\varphi$ with another expression whose set of models is either a subset or superset of those of $\varphi$. Equivalently, the upper bound is a formula $\varphi_U$ such that $\varphi \Rightarrow \varphi_U$, and the lower bound is $\varphi_L$ such that $\varphi_L \Rightarrow \varphi$. We show in this paper, that if $\varphi$ is a positive Boolean formula, then all upper and lower model-based bounds can be obtained by repeated dissociation: the model-based upper bound is obtained by repeatedly setting probabilities of dissociated variables to 1, and the model-based lower bound by setting the probabilities to 0. While the thus generated model-based upper bounds for conjunctive expressions correspond to optimal oblivious dissociation bounds, the model-based lower bounds for conjunctive expressions are *not optimal* and can always be improved by dissociation.

Indeed, consider first the upper bound for conjunctions: the implication $\varphi \Rightarrow \varphi_U$ holds iff there exists a formula $\varphi_1$ such that $\varphi \equiv \varphi_1 \wedge \varphi_U$.[4] Pick a variable $x$, denote $p = \mathbb{P}[x]$ its probability, dissociate it into $x'_1$ in $\varphi_1$ and $x'_2$ in $\varphi_U$, and set their probabilities as $p'_1 = 1$ and $p'_2 = p$. Thus, $\varphi_U$ remains unchanged (except for the renaming of $x$ to $x'_2$), while in $\varphi_1$ we have set $x_1 = 1$. By repeating this process, we eventually transform $\varphi_1$ into true (Recall that our formula is monotone). Thus, model-based upper bounds are obtained by repeated dissociation and setting $p'_1 = 1$ and $p'_2 = p$ at each step. Our results show that this is only one of many oblivious upper bounds as any choices with $p'_1 p'_2 \geq p$ lead to an oblivious upper bound for conjunctive dissociations.

Consider now the lower bound: the implication $\varphi_L \Rightarrow \varphi$ is equivalent to $\varphi_L \equiv \varphi \wedge \varphi_2$. Then there is a sequence of finitely many conjunctive dissociation steps, which transforms $\varphi$ into $\varphi \wedge \varphi_2$ and thus into $\varphi_L$. At each step, a variable $x$ is dissociated into $x'_1$ and $x'_2$, and their probabilities are set to $p'_1 = p$ and $p'_2 = 0$, respectively.[5] According to our result, this choice is not optimal: instead one obtains a tighter bound by also setting $p'_2 = p$, which no longer corresponds to a model-based lower bound.

---

[4] Fink and Olteanu [2011] describe their approach for approximating DNF expressions only. However, the idea of model-based bounds applies equally well to arbitrary Boolean expressions, including those in CNF.
[5] The details here are more involved and are given in detail in Sect. 5.2.

Thus, *model-based lower bounds for conjunctive expressions are not optimal* and can always be improved by using dissociation.

Our dual result states the following for the case when the two formulas are connected with disjunction $\vee$ instead of conjunction $\wedge$: (1) the dissociation is an upper bound iff $p_1' \geq p$ and $p_2' \geq p$, and (2) it is a lower bound iff $(1 - p_1')(1 - p_2') \geq 1 - p$. We see that model-based approximation gives an optimal lower bound for disjunctions, because $(1 - p_1')(1 - p_2') = 1 \cdot (1 - p) = 1 - p$, however non-optimal upper bounds. Example 7.2 illustrates this asymmetry and the possible improvement through dissociation with a detailed simulation-based example.

**Bounds for hard probabilistic queries.** Query evaluation on probabilistic databases reduces to the problem of computing the probability of its lineage expression which is a a monotone, $k$-partite Boolean DNF where $k$ is fixed by the number of joins in the query. Computing the probability of the lineage is known to be #P-hard for some queries [Dalvi and Suciu 2007], hence we are interested in approximating these probabilities by computing dissociated Boolean expressions for the lineage. We have previously shown in [Gatterbauer et al. 2010] that every query plan for a query corresponds to one possible dissociation for its lineage expression. The results in this paper show how to best set the probabilities for the dissociated expressions in order to obtain both upper bounds and lower bounds. We further show that *all the computation* can be pushed inside a standard relational database engine with the help of SQL queries that use User-Defined-Aggregates and views that replace the probabilities of input tuples with their optimal symmetric lower bounds. We illustrate this approach in Sect. 6 and validate it on TPC-H data in Sect. 7.5.

**Main contributions.** (1) We introduce an algebraic framework for approximating the probability of Boolean functions by treating multiple occurrences of variables as independent and assigning them new individual probabilities. We call this approach *dissociation*; (2) we determine the optimal upper and lower bounds for conjunctive and disjunctive dissociations under the assumption of *oblivious* value assignments; (3) we show how existing relaxation-based and model-based approximations fit into the larger design space of dissociations, and show that for every model-based bound there is at least one dissociation-based bound which is as good or tighter; (4) we apply our general framework to both upper *and lower bound* hard probabilistic conjunctive queries without self-joins in guaranteed PTIME by translating the query into a sequence of standard SQL queries; and (5) we illustrate and evaluate with several detailed examples the application of this technique. Note that this paper does *not* address the algorithmic complexities in determining alternative dissociations, in general.

**Outline.** Section 2 starts with some notational background, and Sect. 3 formally defines dissociation. Section 4 contains our main results on optimal oblivious bounds. Section 5 formalizes the connection between relaxation, model-based bounds and dissociation, and shows how both previous approaches can be unified under the framework of dissociation. Section 6 applies our framework to derive upper and lower bounds for hard probabilistic queries with standard relational database management systems. Section 7 gives detailed illustrations on the application of dissociation and oblivious bounds. Finally, Sect. 8 relates to previous work before Sect. 9 concludes.

## 2. GENERAL NOTATIONS AND CONVENTIONS

We use $[m]$ as short notation for $\{1, \ldots, m\}$, use the bar sign for the complement of an event or probability (e.g., $\bar{x} = \neg x$, and $\bar{p} = 1 - p$), and use a bold notation for sets (e.g., $\mathbf{s} \subseteq [m]$) or vectors (e.g., $\mathbf{x} = \langle x_1, \ldots, x_m \rangle$) alike. We assume a set $\mathbf{x}$ of independent Boolean random variables, and assign to each variable $x_i$ a primitive event which is true with probability $p_i = \mathbb{P}[x_i]$. We do not formally distinguish between the variable $x_i$ and the event $x_i$ that it is true. By default, all primitive events are assumed to be independent (e.g., $\mathbb{P}[x_1 \wedge x_2] = p_1 p_2$). We are interested in bounding the probability $\mathbb{P}[f]$ of a Boolean function $f$, i.e. the probability that the function is true if each of the variables is independently true or false with given probabilities. When no confusion arises, we blur the distinction between a Boolean expression $\varphi$ and the Boolean function $f_\varphi$ it represents (cf. [Crama and Hammer 2011, Sect. 1.2]) and write $\mathbb{P}[\varphi]$ instead of $\mathbb{P}[f_\varphi]$. We also use the words formula and expression interchangeably. We write $f(\mathbf{x})$ to indicate that $\mathbf{x}$ is the set of primitive events appearing in the function $f$, and $f[\mathbf{x}_1/\mathbf{x}]$ to indicate that $\mathbf{x}_1$ is substituted for $\mathbf{x}$ in $f$. We often omit the operator $\wedge$ and denote conjunction by mere juxtaposition instead.

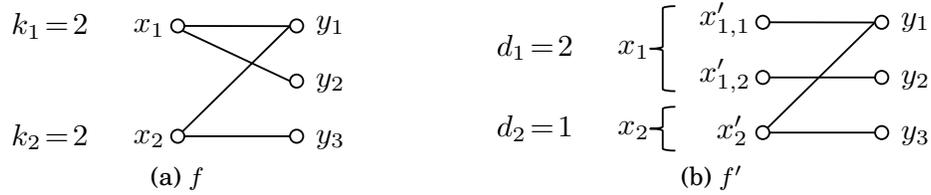

Fig. 2. Example 3.2. (a): Bipartite primal graph for CNF representing $f$. (b): A dissociation $f'$ where variable $x_1$ appearing $k_1 = 2$ times in $f$ is dissociated into (replaced by) $d_1 = 2$ fresh variables in $f'$.

## 3. DISSOCIATION OF BOOLEAN FUNCTIONS AND EXPRESSIONS

We define here dissociation formally. Let $f(\mathbf{x}, \mathbf{y})$ and $f'(\mathbf{x}', \mathbf{y})$ be two Boolean functions, where $\mathbf{x}, \mathbf{x}', \mathbf{y}$ are three disjoint sets of variables. Denote $|\mathbf{x}| = m$, $|\mathbf{x}'| = m'$, and $|\mathbf{y}| = n$. We restrict $f$ and $f'$ to be *positive* in $\mathbf{x}$ and $\mathbf{x}'$, respectively [Crama and Hammer 2011, Def. 1.24].

*Definition* 3.1 (*Dissociation*). We call a function $f'$ a dissociation of $f$ if there exists a substitution $\theta : \mathbf{x}' \rightarrow \mathbf{x}$ s.t. $f'[\theta] = f$.

*Example* 3.2 (*CNF Dissociation*). Consider two functions $f$ and $f'$ given by CNF expressions

$$f = (x_1 \vee y_1)(x_1 \vee y_2)(x_2 \vee y_1)(x_2 \vee y_3)$$
$$f' = (x'_{1,1} \vee y_1)(x'_{1,2} \vee y_2)(x'_2 \vee y_1)(x'_2 \vee y_3)$$

Then $f'$ is a dissociation of $f$ as $f'[\theta] = f$ for the substitution $\theta = \{(x'_{1,1}, x_1), (x'_{1,2}, x_1), (x'_2, x_2)\}$. Figure 2 shows the CNF expressions' primal graphs.[6] ∎

In practice, to find a dissociation for a function $f(\mathbf{x}, \mathbf{y})$, one proceeds like this: Choose any expression $\varphi(\mathbf{x}, \mathbf{y})$ for $f$ and thus $f = f_\varphi$. Replace the $k_i$ distinct occurrences of variables $x_i$ in $\varphi$ with $d_i$ fresh variables $x'_{i,1}, x'_{i,2}, \ldots, x'_{i,d_i}$, with $d_i \leq k_i$. The resulting expression $\varphi'$ represents a function $f'$ that is a dissociation of $f$. Notice that we may obtain different dissociations by deciding for which occurrences of $x_i$ to use distinct fresh variables, and for which occurrences to use the same variable. We may further obtain more dissociations by starting with different, equivalent expressions $\varphi$ for the function $f$. In fact, we may construct infinitely many dissociations this way. We also note that every dissociation of $f$ can be obtained through the process outlined here. Indeed, let $f'(\mathbf{x}', \mathbf{y})$ be a dissociation of $f(\mathbf{x}, \mathbf{y})$ according to Definition 3.1, and let $\theta$ be the substitution for which $f'[\theta] = f$. Then, if $\varphi'$ is any expression representing $f'$, the expression $\varphi = \varphi'[\theta]$ represents $f$. We can thus apply the described dissociation process to a certain expression $\varphi$ and obtain $f'$.

---

[6]The *primal graph* of a CNF (DNF) has one node for each variable and one edge for each pair of variables that co-occur in some clause (conjunct). This concept originates in constraint satisfaction and it is also varyingly called co-occurrence graph or variable interaction graph [Crama and Hammer 2011].

*Example* 3.3 (*Alternative Dissociations*). Consider the two expressions:

$$\varphi = (x \vee y_1)(x \vee y_2)(x \vee y_3)(y_4 \vee y_5)$$
$$\psi = xy_4 \vee xy_5 \vee y_1 y_2 y_3 y_4 \vee y_1 y_2 y_3 y_5$$

Both are equivalent ($\varphi \equiv \psi$) and thus represent the same Boolean function ($f_\varphi = f_\psi$). Yet each leads to a quite different dissociation in the variable $x$:

$$\varphi' = (x_1' \vee y_1)(x_2' \vee y_2)(x_3' \vee y_3)(y_4 \vee y_5)$$
$$\psi' = x_1' y_4 \vee x_2' y_5 \vee y_1 y_2 y_3 y_4 \vee y_1 y_2 y_3 y_5$$

Here, $\varphi'$ and $\psi'$ represent different functions ($f_{\varphi'} \neq f_{\psi'}$) and are both dissociations of $f$ for the substitutions $\theta_1 = \{(x_1', x), (x_2', x), (x_3', x)\}$ and $\theta_2 = \{(x_1', x), (x_2', x)\}$, respectively. ∎

*Example* 3.4 (*More alternative Dissociations*). Consider the AND-function $f(x, y) = xy$. It can be represented by the expressions $xxy$, or $xxxy$, etc., leading to the dissociations $x_1' x_2' y$, or $x_1' x_2' x_3' y$, etc. For even more dissociations, represent $f$ using the expression $(x \vee x)y \vee xy$, which can dissociate to $(x_1' \vee x_2')y \vee x_3' y$, or $(x_1' \vee x_2')y \vee x_1' y$, etc. Note that several occurrences of a variable can be replaced by the same new variables in the dissociated expression. ∎

## 4. OBLIVIOUS BOUNDS FOR DISSOCIATED EVENT EXPRESSIONS

Throughout this section, we fix two Boolean functions $f(\mathbf{x}, \mathbf{y})$ and $f'(\mathbf{x}', \mathbf{y})$ such that $f'$ is a dissociation of $f$. We are given the probabilities $\mathbf{p} = \mathbb{P}[\mathbf{x}]$ and $\mathbf{q} = \mathbb{P}[\mathbf{y}]$. Our goal is to find probabilities $\mathbf{p}' = \mathbb{P}[\mathbf{x}']$ of the dissociated variables so that $\mathbb{P}[f']$ is an upper or lower bound for $\mathbb{P}[f]$. We first define oblivious bounds (Sect. 4.1), then characterize them, in general, through valuations (Sect. 4.2) and, in particular, for conjunctive and disjunctive dissociations (Sect. 4.3), then derive optimal bounds (Sect. 4.4), and end with illustrated examples for CNF and DNF dissociations (Sect. 4.5).

### 4.1. Definition of Oblivious Bounds

We use the subscript notation $\mathbb{P}_{\mathbf{p},\mathbf{q}}[f]$ and $\mathbb{P}_{\mathbf{p}',\mathbf{q}}[f']$ to emphasize that the probability space is defined by the probabilities $\mathbf{p} = \langle p_1, p_2, \ldots \rangle$, $\mathbf{q} = \langle q_1, q_2, \ldots \rangle$, and $\mathbf{p}' = \langle p_1', p_2', \ldots \rangle$, respectively. Given $\mathbf{p}$ and $\mathbf{q}$, our goal is thus to find $\mathbf{p}'$ such that $\mathbb{P}_{\mathbf{p}',\mathbf{q}}[f'] \geq \mathbb{P}_{\mathbf{p},\mathbf{q}}[f]$ or $\mathbb{P}_{\mathbf{p}',\mathbf{q}}[f'] \leq \mathbb{P}_{\mathbf{p},\mathbf{q}}[f]$.

*Definition* 4.1 (*Oblivious Bounds*). Let $f'$ be a dissociation of $f$ and $\mathbf{p} = \mathbb{P}[\mathbf{x}]$. We call $\mathbf{p}'$ an *oblivious upper bound* for $\mathbf{p}$ and dissociation $f'$ of $f$ iff $\forall \mathbf{q} : \mathbb{P}_{\mathbf{p}',\mathbf{q}}[f'] \geq \mathbb{P}_{\mathbf{p},\mathbf{q}}[f]$. Similarly, $\mathbf{p}'$ is an *oblivious lower bound* iff $\forall \mathbf{q} : \mathbb{P}_{\mathbf{p}',\mathbf{q}}[f'] \leq \mathbb{P}_{\mathbf{p},\mathbf{q}}[f]$.

In other words, $\mathbf{p}'$ is an oblivious upper bound if the probability of the dissociated function $f'$ is bigger than that of $f$ *for every choice of* $\mathbf{q}$. Put differently, the probabilities of $\mathbf{x}'$ depend only on the probabilities of $\mathbf{x}$ and not on those of $\mathbf{y}$.

An immediate upper bound is given by $\mathbf{p}' = \mathbf{1}$, since $f$ is monotone and $f'[\mathbf{1}/\mathbf{x}'] = f[\mathbf{1}/\mathbf{x}]$. Similarly, $\mathbf{p} = \mathbf{0}$ is a naïve lower bound. This proves that the set of upper and lower bounds is never empty. Our next goal is to characterize all oblivious bounds and to then find optimal ones.

### 4.2. Characterization of Oblivious Bounds through Valuations

We will give a necessary and sufficient characterization of oblivious bounds, but first we need to introduce some notations. If $f(\mathbf{x}, \mathbf{y})$ is a Boolean function, let $\nu : \mathbf{y} \to \{0, 1\}$ be a truth assignment or valuation for $\mathbf{y}$. We use $\mathbf{v}$ for the vector $\langle \nu(y_1), \ldots, \nu(y_n) \rangle$, and denote with $f[\nu]$ the Boolean function obtained after applying the substitution $\nu$. Note that $f[\nu]$ depends on variables $\mathbf{x}$ only. Furthermore, let $\mathbf{g}$ be $n$ Boolean functions, over variables $\mathbf{z}$. We denote with $\mathbf{g}^\nu$ the Boolean function $\mathbf{g}^\nu = \bigwedge_j g_j^\nu$, where $g_j^\nu = \bar{g}_j$ if $\nu(y_j) = 0$ and $g_j^\nu = g_j$ if $\nu(y_j) = 1$.

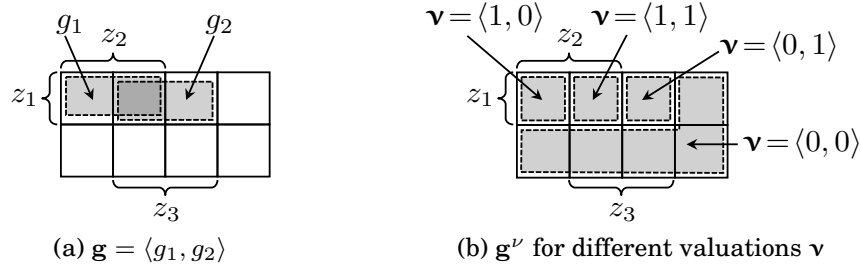

Fig. 3. Example 4.2. Illustration of the valuation notation with Karnaugh maps. (a): Boolean functions $g_1 = z_1 z_2$ and $g_2 = z_1 z_3$. (b): Boolean functions $\mathbf{g}^\nu$ for all 4 possible valuations $\nu$. For example, $\mathbf{g}^\nu = z_1 \bar{z}_2 z_3$ for $\nu = \langle 0, 1 \rangle$.

*Example* 4.2 (*Valuation Notation*). Assume $\mathbf{g} = \langle g_1, g_2 \rangle$ with $g_1 = z_1 z_2$ and $g_2 = z_1 z_3$, and $\nu = \langle 0, 1 \rangle$. Then $\mathbf{g}^\nu = \neg(z_1 z_2) \wedge z_1 z_3 = z_1 \bar{z}_2 z_3$. Figure 3 illustrates our notation for this simple example with the help of Karnaugh maps. We encourage the reader to take a moment and carefully study the correspondences between $\mathbf{g}$, $\nu$, and $\mathbf{g}^\nu$. ∎

Then, any function $f(\mathbf{x}, \mathbf{y})$ admits the following expansion by the $\mathbf{y}$-variables:

$$f(\mathbf{x}, \mathbf{y}) = \bigvee_\nu \left( f[\nu] \wedge \mathbf{y}^\nu \right) \tag{3}$$

Note that any two expressions in the expansion above are logically contradictory, a property called determinism by Darwiche and Marquis [2002], and that it can be seen as the result of applying Shannon's expansion to all variables of $\mathbf{y}$.

*Example* 4.3 (*Valuation Notation continued*). Consider the function $f = (x \vee y_1)(x \vee y_2)$. For the example valuation $\nu = \langle 0, 1 \rangle$, we have $f[\nu] = (x \vee 0)(x \vee 1) = x$ and $\mathbf{y}^\nu = \bar{y}_1 y_2$. Equation 3 gives us an alternative way to write $f$ as disjunction over all $2^2$ valuations of $\mathbf{y}$ as $f = x(\bar{y}_1 \bar{y}_2) \vee x(y_1 \bar{y}_2) \vee x(\bar{y}_1 y_2) \vee y_1 y_2$. ∎

The following proposition is a necessary and sufficient condition for oblivious upper and lower bounds, based on valuations.

PROPOSITION 4.4 (OBLIVIOUS BOUNDS AND VALUATIONS). *Fix two Boolean functions $f(\mathbf{x}, \mathbf{y})$, $f'(\mathbf{x}', \mathbf{y})$ s.t. $f'$ is a dissociation of $f$, and let $\mathbf{p}$ and $\mathbf{p}'$ denote the probabilities of the variables $\mathbf{x}$ and $\mathbf{x}'$, respectively. Then $\mathbf{p}'$ is an oblivious upper bound iff $\mathbb{P}_{\mathbf{p}'}[f'[\nu]] \geq \mathbb{P}_{\mathbf{p}}[f[\nu]]$ for every valuation $\nu$ for $\mathbf{y}$. The proposition holds similarly for oblivious lower bounds.*

PROOF. Remember that any two events in Eq. 3 are disjoint. The total probability theorem thus allows us to sum over the probabilities of all conjuncts:

$$\mathbb{P}_{\mathbf{p}, \mathbf{q}}[f(\mathbf{x}, \mathbf{y})] = \sum_\nu \left( \mathbb{P}_{\mathbf{p}}[f[\nu]] \cdot \mathbb{P}_{\mathbf{q}}[\mathbf{y}^\nu] \right)$$

$$\mathbb{P}_{\mathbf{p}', \mathbf{q}}[f'(\mathbf{x}, \mathbf{y})] = \sum_\nu \left( \mathbb{P}_{\mathbf{p}'}[f'[\nu]] \cdot \mathbb{P}_{\mathbf{q}}[\mathbf{y}^\nu] \right)$$

The "if" direction follows immediately. For the "only if" direction, assume that $\mathbf{p}'$ is an oblivious upper bound. By definition, $\mathbb{P}_{\mathbf{p}', \mathbf{q}}[f'] \geq \mathbb{P}_{\mathbf{p}, \mathbf{q}}[f]$ for every $\mathbf{q}$. Fix any valuation $\nu : \mathbf{y} \to \{0, 1\}$, and define the following probabilities $\mathbf{q}$: $q_i = 0$ when $\nu(y_i) = 0$, and $q_i = 1$ when $\nu(y_i) = 1$. It is easy to see that $\mathbb{P}_{\mathbf{p}, \mathbf{q}}[f] = \mathbb{P}_{\mathbf{p}}[f[\nu]]$, and similarly, $\mathbb{P}_{\mathbf{p}', \mathbf{q}}[f'] = \mathbb{P}_{\mathbf{p}'}[f'[\nu]]$, which proves $\mathbb{P}_{\mathbf{p}'}[f'[\nu]] \geq \mathbb{P}_{\mathbf{p}}[f[\nu]]$. □

A consequence of choosing $\mathbf{p}'$ obliviously is that it remains a bound even if we allow the variables $\mathbf{y}$ to be arbitrarily correlated. More precisely:

COROLLARY 4.5 (OBLIVIOUS BOUNDS AND CORRELATIONS). *Let $f'(\mathbf{x}', \mathbf{y})$ be a dissociation of $f(\mathbf{x}, \mathbf{y})$, let $\mathbf{p}'$ be an oblivious upper bound for $\mathbf{p}$, and let $\mathbf{g} = \langle g_1, \ldots, g_{|\mathbf{y}|} \rangle$ be Boolean functions in some variables $\mathbf{z}$ with probabilities $\mathbf{r} = \mathbb{P}[\mathbf{z}]$. Then: $\mathbb{P}_{\mathbf{p}',\mathbf{r}}[f'(\mathbf{x}', \mathbf{g}(\mathbf{z}))] \geq \mathbb{P}_{\mathbf{p},\mathbf{r}}[f(\mathbf{x}, \mathbf{g}(\mathbf{z}))]$. The result for oblivious lower bounds is similar.*

The intuition is that, by substituting the variables $\mathbf{y}$ with functions $\mathbf{g}$ in $f(\mathbf{x}, \mathbf{y})$, we make $\mathbf{y}$ correlated. The corollary thus says that an oblivious upper bound remains an upper bound even if the variables $\mathbf{y}$ are correlated. This follows from folklore that any correlation between the variables $\mathbf{y}$ can be captured by general Boolean functions $\mathbf{g}$. For completeness, we include the proof in Appendix B.

PROOF OF COROLLARY 4.5. We derive the probabilities of $f$ and $f'$ from Eq. 3:

$$\mathbb{P}_{\mathbf{p},\mathbf{r}}[f(\mathbf{x}, \mathbf{g})] = \sum_{\nu} \left( \mathbb{P}_{\mathbf{p}}[f[\nu]] \cdot \mathbb{P}_{\mathbf{r}}[\mathbf{g}^{\nu}] \right)$$

$$\mathbb{P}_{\mathbf{p}',\mathbf{r}}[f'(\mathbf{x}, \mathbf{g})] = \sum_{\nu} \left( \mathbb{P}_{\mathbf{p}'}[f'[\nu]] \cdot \mathbb{P}_{\mathbf{r}}[\mathbf{g}^{\nu}] \right)$$

The proof follows now immediately from Prop. 4.4.  □

### 4.3. Oblivious Bounds for Unary Conjunctive and Disjunctive Dissociations

A dissociation $f'(\mathbf{x}', \mathbf{y})$ of $f(\mathbf{x}, \mathbf{y})$ is called *unary* if $|\mathbf{x}| = 1$, in which case we write the function as $f(x, \mathbf{y})$. We next focus on unary dissociations, and establish a necessary and sufficient condition for probabilities to be oblivious upper or lower bounds for the important classes of conjunctive and disjunctive dissociations. The criterion also extends as a sufficient condition to non-unary dissociations, since these can be obtained as a sequence of unary dissociations.[7]

*Definition* 4.6 (*Conjunctive and Disjunctive Dissociation*). Let $f'(\mathbf{x}', \mathbf{y})$ be a Boolean function in variables $\mathbf{x}', \mathbf{y}$. We say that the variables $\mathbf{x}'$ are *conjunctive* in $f'$ if $f'(\mathbf{x}', \mathbf{y}) = \bigwedge_{j \in [d]} f_j(x'_j, \mathbf{y})$, $d = |\mathbf{x}'|$. We say that a dissociation $f'(\mathbf{x}', \mathbf{y})$ of $f(x, \mathbf{y})$ is conjunctive if $\mathbf{x}'$ are conjunctive in $f'$. Similarly, we say that $\mathbf{x}'$ are *disjunctive* in $f'$ if $f'(\mathbf{x}', \mathbf{y}) = \bigvee_{j \in [d]} f_j(x'_j, \mathbf{y})$, and a dissociation is disjunctive if $\mathbf{x}'$ is disjunctive in $f'$.

Thus in a conjunctive dissociation, each dissociated variable $x'_j$ occurs in exactly one Boolean function $f_j$ and these functions are combined by $\wedge$ to obtain $f'$. In practice, we start with $f$ written as a conjunction, then replace $x$ with a fresh variable in each conjunct:

$$f(x, \mathbf{y}) = \bigwedge f_j(x, \mathbf{y})$$

$$f'(\mathbf{x}', \mathbf{y}) = \bigwedge f_j(x'_j, \mathbf{y})$$

Disjunctive dissociations are similar.

Note that if $\mathbf{x}'$ is conjunctive in $f'(\mathbf{x}', \mathbf{y})$, then for any substitution $\nu : \mathbf{y} \to \{0, 1\}$, $f'[\nu]$ is either 0, 1, or a conjunction of variables in $\mathbf{x}'$: $f'[\nu] = \bigwedge_{j \in \mathbf{s}} x'_j$, for some set $\mathbf{s} \subseteq [d]$, where $d = |\mathbf{x}'|$. Similarly, if $\mathbf{x}'$ is disjunctive, then $f'[\nu]$ is 0, 1, or $\bigvee_{j \in \mathbf{s}} x'_j$.[8]

---

[7]Necessity does not always extend to non-unary dissociations. The reason is that an oblivious dissociation for $\mathbf{x}$ may set the probability of a fresh variable by examining *all* variables $\mathbf{x}$, while in a sequence of oblivious dissociations each new probability $\mathbb{P}[x'_{i,j}]$ may depend only on the variable $x_i$ currently being dissociated.

[8] Note that for $\mathbf{s} = \emptyset$: $f'[\nu] = \bigwedge_{j \in \mathbf{s}} x'_j = 1$ and $f'[\nu] = \bigvee_{j \in \mathbf{s}} x'_j = 0$.

We need one more definition before we state the main result in our paper.

*Definition* 4.7 (*Cover*). Let $\mathbf{x}'$ be conjunctive in $f'(\mathbf{x}', \mathbf{y})$. We say that $f'$ *covers the set* $\mathbf{s} \subseteq [d]$ if there exists a substitution $\nu$ s.t. $f'[\nu] = \bigwedge_{j \in \mathbf{s}} x'_j$. Similarly, if $\mathbf{x}'$ is disjunctive, then we say that $f'$ covers s if there exists $\nu$ s.t. $f'[\nu] = \bigvee_{j \in \mathbf{s}} x'_j$.

THEOREM 4.8 (OBLIVIOUS BOUNDS). *Let $f'(\mathbf{x}', \mathbf{y})$ be a conjunctive dissociation of $f(x, \mathbf{y})$, and let $p = \mathbb{P}[x]$, $\mathbf{p}' = \mathbb{P}[\mathbf{x}']$ be probabilities of $x$ and $\mathbf{x}'$, respectively. Then:*

(1) *If $p'_j \leq p$ for all $j$, then $\mathbf{p}'$ is an oblivious lower bound for $p$, i.e. $\forall \mathbf{q} : \mathbb{P}_{\mathbf{p}', \mathbf{q}}[f'] \leq \mathbb{P}_{p, \mathbf{q}}[f]$. Conversely, if $\mathbf{p}'$ is an oblivious lower bound for $p$ and $f'$ covers all singleton sets $\{j\}$ with $j \in [d]$, then $p'_j \leq p$ for all $j$.*

(2) *If $\prod_j p'_j \geq p$, then $\mathbf{p}'$ is an oblivious upper bound for $p$, i.e. $\forall \mathbf{q} : \mathbb{P}_{\mathbf{p}', \mathbf{q}}[f'] \geq \mathbb{P}_{p, \mathbf{q}}[f]$. Conversely, if $\mathbf{p}'$ is an oblivious upper bound for $p$ and $f'$ covers the set $[d]$, then $\prod_j p'_j \geq p$.*

*Similarly, the dual result holds for disjunctive dissociations $f'(\mathbf{x}', \mathbf{y})$ of $f(x, \mathbf{y})$:*

(3) *If $p'_j \geq p$ for all $j$, then $\mathbf{p}'$ is an oblivious upper bound for $p$. Conversely, if $\mathbf{p}'$ is an oblivious upper bound for $p$ and $f'$ covers all singleton sets $\{j\}$, $j \in [d]$, then $p'_j \geq p$.*

(4) *If $\prod_j (1 - p'_j) \geq 1 - p$, then $\mathbf{p}'$ is an oblivious lower bound for $p$. Conversely, if $\mathbf{p}'$ is an oblivious lower bound for $p$ and $f'$ covers the set $[d]$, then $\prod_j (1 - p'_j) \geq 1 - p$.*

PROOF. We make repeated use of Prop. 4.4. We give here the proof for conjunctive dissociations only; the proof for disjunctive dissociations is dual and similar.

(1) We need to check $\mathbb{P}_{\mathbf{p}'}[f'[\nu]] \leq \mathbb{P}_p[f[\nu]]$ for every $\nu$. Since the dissociation is unary, $f[\nu]$ can be only 0, 1, or $x$, while $f'[\nu]$ is 0, 1, or $\bigwedge_{j \in \mathbf{s}} x'_j$ for some set $\mathbf{s} \subseteq [d]$.

Case 1: $f[\nu] = 0$. We will show that $f'[\nu] = 0$, which implies $\mathbb{P}_{\mathbf{p}'}[f'[\nu]] = \mathbb{P}_p[f[\nu]] = 0$. Recall that, by definition, $f'(\mathbf{x}', \mathbf{y})$ becomes $f(x, \mathbf{y})$ if we substitute $x$ for all variables $x'_j$. Therefore, $f'[\nu][x/x'_1, \ldots, x/x'_d] = 0$, which implies $f'[\nu] = 0$ because $f'$ is monotone in the variables $\mathbf{x}'$.

Case 2: $f[\nu] = 1$. Then $\mathbb{P}_{\mathbf{p}'}[f'[\nu]] \leq \mathbb{P}_p[f[\nu]]$ holds trivially.

Case 3: $f[\nu] = x$. Then $\mathbb{P}_p[f[\nu]] = p$, while $\mathbb{P}_{\mathbf{p}'}[f'[\nu]] = \prod_{j \in \mathbf{s}} p'_j$. We prove that $\mathbf{s} \neq \emptyset$: this implies our claim, because $\prod_{j \in \mathbf{s}} p'_j \leq p'_j \leq p$, for any choice of $j \in \mathbf{s}$. Suppose otherwise, that $\mathbf{s} = \emptyset$, hence $f'[\nu] = 1$. Substituting all variables $\mathbf{x}'$ with $x$ transforms $f'$ to $f$. This implies $f[\nu] = 1$, which contradicts $f[\nu] = x$.

For the converse, assume that $\mathbf{p}'$ is an oblivious lower bound. Since $f'$ covers $\{j\}$, there exists a substitution $\nu$ s.t. $f'[\nu] = x'_j$, and therefore $f[\nu] = x$. By Prop. 4.4 we have $p'_j = \mathbb{P}_{\mathbf{p}'}[f'[\nu]] \leq \mathbb{P}_p[f[\nu]] = p$, proving the claim.

(2) Here we need to check $\mathbb{P}_{\mathbf{p}'}[f'[\nu]] \geq \mathbb{P}_p[f[\nu]]$ for every $\nu$. The cases when $f[\nu]$ is either 0 or 1 are similar to the cases above, so we only consider the case when $f[\nu] = x$. Then $f'[\nu] = \bigwedge_{j \in \mathbf{s}} x'_j$ and $\mathbb{P}_{\mathbf{p}'}[f'[\nu]] = \prod_{j \in \mathbf{s}} p'_j \geq \prod_j p_j \geq p = \mathbb{P}_{\mathbf{p}}[f[\nu]]$.

For the converse, assume $\mathbf{p}'$ is an oblivious upper bound, and let $\nu$ be the substitution for which $f'[\nu] = \bigwedge_j x'_j$ (which exists since $f'$ is covers $[d]$). Then $\mathbb{P}_{\mathbf{p}'}[f'[\nu]] \geq \mathbb{P}_{\mathbf{p}}[f[\nu]]$ implies $p \leq \prod_j p_j$.

□

## 4.4. Optimal Oblivious Bounds for Unary Conjunctive and Disjunctive Dissociations

We are naturally interested in the "best possible" oblivious bounds. Call a dissociation $f'$ *non-degenerate* if it covers all singleton sets $\{j\}$, $j \in [d]$ and the complete set $[d]$. Theorem 4.8 then implies:

Corollary 4.9 (Optimal Oblivious Bounds). *If $f'$ is a conjunctive dissociation of $f$ and $f'$ is non-degenerate, then the optimal oblivious lower bound is $p'_1 = p'_2 = \ldots = p$, while the optimal oblivious upper bounds are obtained whenever $p'_1 p'_2 \cdots = p$. Similarly, if $f'$ is a disjunctive dissociation of $f$ and $f'$ is non-degenerate, then the optimal oblivious upper bound is $p'_1 = p'_2 = \ldots = p$, while the optimal oblivious lower bounds are obtained whenever $(1 - p'_1) \cdot (1 - p'_2) \cdots = 1 - p$.*

Notice that while optimal lower bounds for conjunctive dissociations and optimal upper bounds for disjunctive dissociations are uniquely defined with $p'_j = p$, there are infinitely many optimal bounds for the other directions (see Fig. 1). Let us call bounds *symmetric* if all dissociated variable have the same probability. Then optimal symmetric upper bounds for conjunctive dissociations are $p'_j = \sqrt[d]{p}$, and optimal symmetric lower bounds for disjunctive dissociations $p'_j = 1 - \sqrt[d]{1 - p}$.

We give two examples of degenerate dissociations. First, the dissociation $f' = (x'_1 y_1 \vee y_3) \wedge (x'_2 y_2 \vee y_3)$ does not cover either $\{1\}$ nor $\{2\}$: no matter how we substitute $y_1, y_2, y_3$, we can never transform $f'$ to $x'_1$. For example, $f'[1/y_1, 0/y_2, 0/y_3] = 0$ and $f'[1/y_1, 0/y_2, 1/y_3] = 1$. But $f'$ does cover the set $\{1, 2\}$ because $f'[1/y_1, 1/y_2, 0/y_3] = x_1 x_2$. Second, the dissociation $f' = (x'_1 y_1 \vee y_2) \wedge (x'_2 y_2 \vee y_1)$ covers both $\{1\}$ and $\{2\}$, but does not cover the entire set $\{1, 2\}$. In these cases the oblivious upper or lower bounds in Theorem 4.8 still hold, but are not necessarily optimal.

However, most cases of practical interest result in dissociations that are non-degenerate, in which case the bounds in Theorem 4.8 are tight. We explain this here. Consider the original function, pre-dissociation, written in a conjunctive form:

$$f(x, \mathbf{y}) = g_0 \wedge \bigwedge_{j \in [d]} (x \vee g_j) = g_0 \wedge \bigwedge_{j \in [d]} f_j \qquad (4)$$

where each $g_j$ is a Boolean function in the variables $\mathbf{y}$, and where we denoted $f_j = x \vee g_j$. For example, if $f$ is a CNF expression, then each $f_j$ is a clause containing $x$, and $g_0$ is the conjunction of all clauses that do not contain $x$. Alternatively, we may start with a CNF expression, and group the clauses containing $x$ in equivalence classes, such that each $f_j$ represents one equivalence class. For example, starting with four clauses, we group into two functions $f = [(x \vee y_1)(x \vee y_2)] \wedge [(x \vee y_3)(x \vee y_4)] = (x \vee y_1 y_2) \wedge (x \vee y_3 y_4) = f_1 \wedge f_2$. Our only assumption about Eq. 4 is that it is *non-redundant*, meaning that none of the expressions $g_0$ or $f_j$ may be dropped. Then we prove:

Proposition 4.10 (Non-degenerate Dissociation). *Suppose the function $f$ in Eq. 4 is non-redundant. Define $f' = g_0 \wedge \bigwedge_j (x'_j \vee g_j)$. Then $f'$ covers every singleton set $\{j\}$. Moreover, if the implication $g_0 \Rightarrow \bigvee_j g_j$ does not hold, then $f'$ also covers the set $[d]$. Hence $f'$ is non-degenerate. A similar result holds for disjunctive dissociations if the dual implication $g_0 \Leftarrow \bigwedge_j g_j$ does not hold.*

Proof. We give here the proof for conjunctive dissociations only; the proof for disjunctive dissociations follows from duality. We first prove that $f'$ covers any singleton set $\{j\}$, for $j \in [d]$. We claim that the following logical implication does not hold:

$$g_0 \wedge \bigwedge_{i \neq j} g_i \Rightarrow g_j \qquad (5)$$

Indeed, suppose the implication holds for some $j$. Then the following implication also holds: $g_0 \wedge \bigwedge_{i \neq j} (x \vee g_i) \Rightarrow (x \vee g_j)$, since for $x = 0$ it is the implication above, while for $x = 1$ it is a tautology. Therefore, the function $f_j$ is redundant in Eq. 4, which contradicts our assumption. Hence, the implication Eq. 5 does not hold. Let $\nu$ be any

| Valuation | | CNF dissociation | | | | DNF dissociation | | | |
|---|---|---|---|---|---|---|---|---|---|
| # | $\nu$ | $f'_c[\nu]$ | $f_c[\nu]$ | $\mathbb{P}[f'_c[\nu]]$ | $\geq \mathbb{P}[f_c[\nu]]$ | $f'_d[\nu]$ | $f_d[\nu]$ | $\mathbb{P}[f'_d[\nu]]$ | $\geq \mathbb{P}[f_d[\nu]]$ |
| 1. | $\langle 0,0,0,0 \rangle$ | 0 | 0 | 0 | $\geq$ 0 | 0 | 0 | 0 | $\geq$ 0 |
| 2. | $\langle 1,0,0,0 \rangle$ | 0 | 0 | 0 | $\geq$ 0 | $x'_1$ | $x$ | $p'_1$ | $\geq$ $p$ |
| 3. | $\langle 0,1,0,0 \rangle$ | 0 | 0 | 0 | $\geq$ 0 | $x'_2$ | $x$ | $p'_2$ | $\geq$ $p$ |
| 4. | $\langle 0,0,1,0 \rangle$ | 0 | 0 | 0 | $\geq$ 0 | $x'_3$ | $x$ | $p'_3$ | $\geq$ $p$ |
| 5. | $\langle 0,0,0,1 \rangle$ | $x'_1 x'_2 x'_3$ | $x$ | $p'_1 p'_2 p'_3$ | $\geq$ $p$ | 1 | 1 | 1 | $\geq$ 1 |
| 6. | $\langle 1,1,0,0 \rangle$ | 0 | 0 | 0 | $\geq$ 0 | $x'_1 \vee x'_2$ | $x$ | $\bar{p}'_1 \bar{p}'_2$ | $\geq$ $\bar{p}$ |
| 7. | $\langle 1,0,1,0 \rangle$ | 0 | 0 | 0 | $\geq$ 0 | $x'_1 \vee x'_3$ | $x$ | $\bar{p}'_1 \bar{p}'_3$ | $\geq$ $\bar{p}$ |
| 8. | $\langle 0,1,1,0 \rangle$ | 0 | 0 | 0 | $\geq$ 0 | $x'_2 \vee x'_3$ | $x$ | $\bar{p}'_2 \bar{p}'_3$ | $\geq$ $\bar{p}$ |
| 9. | $\langle 1,0,0,1 \rangle$ | $x'_2 x'_3$ | $x$ | $p'_2 p'_3$ | $\geq$ $p$ | 1 | 1 | 1 | $\geq$ 1 |
| 10. | $\langle 0,1,0,1 \rangle$ | $x'_1 x'_3$ | $x$ | $p'_1 p'_3$ | $\geq$ $p$ | 1 | 1 | 1 | $\geq$ 1 |
| 11. | $\langle 0,0,1,1 \rangle$ | $x'_1 x'_2$ | $x$ | $p'_1 p'_2$ | $\geq$ $p$ | 1 | 1 | 1 | $\geq$ 1 |
| 12. | $\langle 1,1,1,0 \rangle$ | 0 | 0 | 0 | $\geq$ 0 | $x'_1 \vee x'_2 \vee x'_3$ | $x$ | $\bar{p}'_1 \bar{p}'_2 \bar{p}'_3$ | $\geq$ $\bar{p}$ |
| 13. | $\langle 1,1,0,1 \rangle$ | $x'_3$ | $x$ | $p'_3$ | $\geq$ $p$ | 1 | 1 | 1 | $\geq$ 1 |
| 14. | $\langle 1,0,1,1 \rangle$ | $x'_2$ | $x$ | $p'_2$ | $\geq$ $p$ | 1 | 1 | 1 | $\geq$ 1 |
| 15. | $\langle 0,1,1,1 \rangle$ | $x'_1$ | $x$ | $p'_1$ | $\geq$ $p$ | 1 | 1 | 1 | $\geq$ 1 |
| 16. | $\langle 1,1,1,1 \rangle$ | 1 | 1 | 1 | $\geq$ 1 | 1 | 1 | 1 | $\geq$ 1 |

(a) Comparing $2^4$ valuations for determining oblivious bounds.

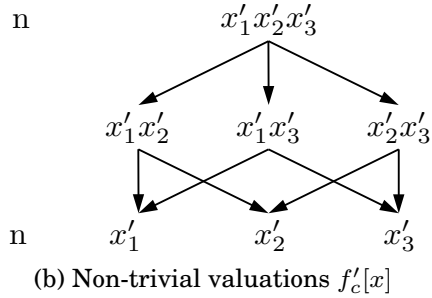

(b) Non-trivial valuations $f'_c[x]$

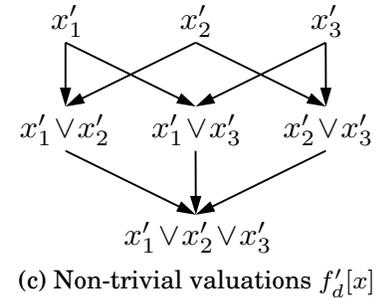

(c) Non-trivial valuations $f'_d[x]$

Fig. 4. Example 4.11 (CNF $f_c$) and Example 4.12 (DNF $f_d$). (a): Determining oblivious bounds by ensuring that bounds hold for all valuations. (b), (c): Partial order of implication ($\Rightarrow$) for the non-trivial valuations $f_c[\nu]'$ and $f_d[\nu]'$, e.g.: from $x'_1 x'_2 \Rightarrow x'_1$ it follows that $p'_1 p'_2 \geq p \Rightarrow p'_1 \geq p$. Note that $f_c \neq f_d$.

assignment that causes Eq. 5 to fail: thus, for all $j \in \{0, \ldots, d\}$, $j \neq i$, $g_i[\nu] = 1$ and $g_j[\nu] = 0$. Therefore $f'[\nu] = x_j$, proving that it covers $\{j\}$.

Next, assume that $g_0 \Rightarrow \bigvee_j g_j$ does not hold. We prove that $f'$ covers $[d]$. Let $\nu$ be any substitution that causes the implication to fail: $g_0[\nu] = 1$ and $g_j[\nu] = 0$ for $j \in [d]$. Then $f'[\nu] = \bigwedge_{j \in [d]} x'_j$. □

### 4.5. Illustrated Examples for Optimal Oblivious Bounds

We next give two examples that illustrate optimal oblivious bounds for conjunctive and disjunctive dissociations in some detail.

*Example* 4.11 (*CNF Dissociation*). Consider the function $f_c$ given by an CNF expression and its dissociation $f'_c$:

$$f_c = (x \vee y_1)(x \vee y_2)(x \vee y_3)y_4$$
$$f'_c = (x'_1 \vee y_1)(x'_2 \vee y_2)(x'_3 \vee y_3)y_4$$

There are $2^4 = 16$ valuations for $\mathbf{y} = \langle y_1, y_2, y_3, y_4 \rangle$. Probabilities $\mathbf{p}' = \langle p'_1, p'_2, p'_3 \rangle$ are thus an oblivious *upper* bound exactly if they satisfy the 16 inequalities given under "CNF dissociation" in Fig. 4a. For valuations with $\nu_4 = 0$ (and thus $f_c[\nu] = 0$) or all $\nu_j = 1$ (and thus $f_c'[\nu] = 1$) the inequalities trivially hold. For the remaining 7 non-trivial inequalities, $p_1' p_2' p_3' \geq p$ implies all others. Figure 4b shows the partial order between the non-trivial valuations, with $x_1' x_2' x_3'$ implying all others. Since $f_c$ and $f_c'$ are positive in x and x′, respectively, it follows that optimal oblivious upper bounds are given by $p_1' p_2' p_3' = p$, e.g., by setting $p_i' = \sqrt[3]{p}$ for symmetric bounds.

Oblivious *lower* bounds are given by the 16 inequalities after inverting the inequality sign. Here we see that the three inequalities $p_j' \leq p$ together imply the others. Hence, oblivious lower bounds are those that satisfy all three inequalities. The only optimal oblivious upper bounds are then given by $p_j' = p$. ∎

*Example* 4.12 (*DNF Dissociation*). Consider the function $f_d$ given by an DNF expression and its dissociation $f_d'$:

$$f_d = x y_1 \lor x y_2 \lor x y_3 \lor y_4$$
$$f_d' = x_1' y_1 \lor x_2' y_2 \lor x_3' y_3 \lor y_4$$

An oblivious upper bound $\mathbf{p}' = \langle p_1', p_2', p_3' \rangle$ must thus satisfy the 16 inequalities[9] given under "DNF dissociation" in Fig. 4a. For valuations with $\nu_4 = 1$ (and thus $f_d'[\nu]$ or $\nu_j = 0$ (and thus $f_d[\nu] = 0$) the inequalities trivially hold. For the remaining inequalities we see that the elements of set $\{x_1', x_2', x_3'\}$ together imply all others, and that $x_1' \lor x_2' \lor x_3'$ is implied by all others (Fig. 4c shows the partial order between the non-trivial valuations). Thus, an oblivious upper bound must satisfy $p_j' \geq p$, and the optimal one is given by $p_j' = p$. Analogously, an oblivious lower bound must satisfy $\bar{p}_1' \bar{p}_2' \bar{p}_3' \geq \bar{p}$. Optimal ones are given for $\bar{p}_1' \bar{p}_2' \bar{p}_3' = \bar{p}$, e.g., by setting $p_j' = 1 - \sqrt[3]{\bar{p}}$. ∎

# 5. RELAXATION AND MODEL-BASED BOUNDS AS DISSOCIATION

This section formalizes the connection between relaxation, model-based bounds and dissociation that was outlined in the introduction. In other words, we show how both previous approaches can be unified under the framework of dissociation.

## 5.1. Relaxation & Compensation

The following proposition shows relaxation & compensation as conjunctive dissociation and was brought to our attention by Choi and Darwiche [2011].

PROPOSITION 5.1 (COMPENSATION AND CONJUNCTIVE DISSOCIATION). *Let* $f_1$, $f_2$ *be two monotone Boolean functions which share only one single variable* $x$. *Let* $f$ *be their conjunction, and* $f'$ *be the dissociation of* $f$ *on* $x$, *i.e.*

$$f = f_1 \land f_2$$
$$f' = f_1[x_1'/x] \land f_2[x_2'/x]$$

*Then* $\mathbb{P}[f] = \mathbb{P}[f']$ *for* $\mathbb{P}[x_1'] = \mathbb{P}[x]$ *and* $\mathbb{P}[x_2'] = \mathbb{P}[x | f_1]$.

PROOF OF PROP. 5.1. First, note that $\mathbb{P}[f] = \mathbb{P}[f_1] \mathbb{P}[f_2 | f_1]$. On the other hand, $\mathbb{P}[f'] = \mathbb{P}[f_1'] \mathbb{P}[f_2']$ as $f_1'$ and $f_2'$ are independent after dissociating on the only shared variable $x$. We also have $\mathbb{P}[f_1] = \mathbb{P}[f_1']$ since $\mathbb{P}[x] = \mathbb{P}[x_1']$. It remains to be shown that

---

[9]Remember that the probability of a disjunction of two independent events is $\mathbb{P}[x_1' \lor x_2'] = 1 - \bar{p}_1' \bar{p}_2'$.

$\mathbb{P}[f_2'] = \mathbb{P}[f_2|f_1]$. Indeed:

$$\begin{aligned}
\mathbb{P}[f_2'] &= \mathbb{P}[x_2' \cdot f_2'[1/x_2'] \vee \bar{x}_2' \cdot f_2'[0/x_2']] \\
&= \mathbb{P}[x_2'] \cdot \mathbb{P}[f_2'[1/x_2']] + \mathbb{P}[\bar{x}_2'] \cdot \mathbb{P}[f_2'[0/x_2']] \\
&= \mathbb{P}[x|f_1] \cdot \mathbb{P}[f_2[1/x]] + \mathbb{P}[\bar{x}|f_1] \cdot \mathbb{P}[f_2[0/x]] \\
&= \mathbb{P}[x|f_1] \cdot \mathbb{P}[f_2[1/x]|f_1] + \mathbb{P}[\bar{x}|f_1] \cdot \mathbb{P}[f_2[0/x]|f_1] \\
&= \mathbb{P}[f_2|f_1]
\end{aligned}$$

which proves the claim.  □

Note that compensation is not oblivious, since the probability $p_2'$ depends on the other variables occurring in $\varphi_1$. Further note that, in general, $\varphi_1, \varphi_2$ have more than one variable in common, and in this case we have $\mathbb{P}[\varphi'] \neq \mathbb{P}[\varphi]$ for the same compensation. Thus in general, compensation is applied as a heuristics, and it is then not known whether it provides an upper or lower bound.

The dual result for disjunctions holds by replacing $f_1$ with its negation $\bar{f}_1$ in $\mathbb{P}[x_2'] = \mathbb{P}[x|\bar{f}_1]$. This result is not immediately obvious from the previous one and has, to our best knowledge, not been stated or applied anywhere before.

PROPOSITION 5.2 ("DISJUNCTIVE COMPENSATION").  *Let $f_1$, $f_2$ be two monotone Boolean functions which share only one single variable $x$. Let $f$ be their disjunction, and $f'$ be the dissociation of $f$ on $x$, i.e. $f = f_1 \vee f_2$, and $f' = f_1[x_1'/x] \vee f_2[x_2'/x]$. Then $\mathbb{P}[f] = \mathbb{P}[f']$ for $\mathbb{P}[x_1'] = \mathbb{P}[x]$ and $\mathbb{P}[x_2'] = \mathbb{P}[x|\bar{f}_1]$.*

PROOF OF PROP. 5.2.  Let $g = \bar{f}$, $g_1 = \bar{f}_1$, and $g_2 = \bar{f}_2$. Then $f = f_1 \vee f_2$ is equivalent to $g = g_1 \wedge g_2$. From Prop. 5.1, we know that $\mathbb{P}[g] = \mathbb{P}[g']$, and thus $\mathbb{P}[f] = \mathbb{P}[f']$, for $\mathbb{P}[x_1'] = \mathbb{P}[x]$ and $\mathbb{P}[x_2'] = \mathbb{P}[x|g_1] = \mathbb{P}[x|\bar{f}_1]$.  □

### 5.2. Model-based Approximation

The following proposition shows that all model-based bounds can be derived by repeated dissociation. However, not all dissociation-bounds can be explained as models since dissociation is in its essence an *algebraic* and not a model-based technique (dissociation creates more variables and thus changes the probability space). Therefore, dissociation can improve any existing model-based approximation approach. Example 7.2 will illustrate this with a detailed simulation-based example.

PROPOSITION 5.3 (MODEL-BASED BOUNDS AS DISSOCIATIONS).  *Let $f$, $f_U$ be two monotone Boolean functions over the same set of variables, and for which the logical implication $f \Rightarrow f_U$ holds. Then: (a) there exists a sequence of optimal conjunctive dissociations that transform $f$ to $f_U$, and (b) there exists a sequence of non-optimal disjunctive dissociations that transform $f$ to $f_U$. The dual result holds for the logical implication $f_L \Rightarrow f$: (c) there exists a sequence of optimal disjunctive dissociations that transform $f$ to $f_L$, and (d) there exists a sequence of non-optimal conjunctive dissociations that transform $f$ to $f_L$.*

PROOF OF PROP. 5.3.  We focus here on the implication $f \Rightarrow f_U$. The proposition for the results $f_L \Rightarrow f$ then follows from duality.

(a) The implication $f \Rightarrow f_U$ holds iff there exists a positive function $f_2$ such that $f = f_U \wedge f_2$. Pick a set of variables $\mathbf{x}$ s.t. $f_2[\mathbf{1/x}] = 1$, and dissociate $f$ on $\mathbf{x}$ into $f' = f_U[\mathbf{x_1'/x}] \wedge f_2[\mathbf{x_2'/x}]$. By setting the probabilities of the dissociated variables to $\mathbf{p_1' = p}$ and $\mathbf{p_2' = 1}$, the bounds become optimal ($p_1'p_2' = p$). Furthermore, $f_U$ remains unchanged (except for the renaming of $\mathbf{x}$ to $\mathbf{x_1'}$), whereas $f_2$ becomes true. Hence, we get $f' = f_U$. Thus, all model-based upper bounds can be obtained by conjunctive dissociation and choosing optimal oblivious bounds at each dissociation step.

(b) The implication $f \Rightarrow f_U$ also holds iff there exists a positive function $f_d$ such that $f_U = f \vee f_d$. Let $m$ be the positive minterm or elementary conjunction involving all variables of $f$. The function $f_d$ can be then written as DNF $f_d = c_1 \vee c_2 \vee \dots$, with products $c_i \subseteq m$. Since $f$ is monotone, we know $m \Rightarrow f$, and thus also $m f_d \Rightarrow f$. We can therefore write $f = f \vee m f_d$ or as

$$f = f \vee m c_1 \vee m c_2 \vee \dots$$

Let $\mathbf{x}_i$ be the set of all variables in $m$ that do not occur in $c_i$ and denote with $m_i$ the conjunction of $\mathbf{x}_i$. Then then each $m c_i$ can instead be written as $m_i c_i$ and thus:

$$f = f \vee m_1 c_1 \vee m_2 c_2 \vee \dots$$

WLOG, we now separate one particular conjunct $m_i c_i$ and dissociate on the set $\mathbf{x}_i$

$$f' = \underbrace{f \vee m_1 c_1 \vee m_2 c_2 \vee \dots}_{f_1}[\mathbf{x}'_1/\mathbf{x}_i] \vee \underbrace{m_i c_i}_{f_2}[\mathbf{x}'_2/\mathbf{x}_i]$$

By setting the probabilities of the dissociated variables to the non-optimal upper bounds $\mathbf{p}'_1 = \mathbf{p}$ and $\mathbf{p}'_2 = \mathbf{1}$, $f_1$ remains unchanged (except for the renaming of $\mathbf{x}_i$ to $\mathbf{x}'_1$), whereas $f_2$ becomes $c_i$. Hence, we get $f' = f \vee m_1 c_1 \vee m_2 c_2 \vee \dots \vee c_i$. We can now repeat the same process for all conjuncts $m c_i$ and receive after a finite number of dissociation steps

$$f'' = f \vee (c_1 \vee c_2 \vee \dots) = f \vee f_d$$

Hence $f'' = f_U$. Thus, all model-based upper bounds can be obtained by disjunctive dissociation and choosing non-optimal bounds at each dissociation step. □

## 6. QUERY-CENTRIC DISSOCIATION BOUNDS FOR PROBABILISTIC QUERIES

Our previous work [Gatterbauer et al. 2010] has shown how to *upper bound* the probability of conjunctive queries without self-joins by issuing a sequence of SQL statements over a standard relational DBMS. This section illustrates such dissociation-based upper bounds and also complements them with *new lower bounds*. We use the Boolean query $Q :\!- R(X), S(X, Y), T(Y)$, for which the probability computation problem is known to be #P-hard, over the following database instance $D$:

| $R$ | $A$ | | $S$ | $A$ | $B$ | | $T$ | $B$ |
|---|---|---|---|---|---|---|---|---|
| $x_1$ | 1 | | $z_1$ | 1 | 1 | | $y_1$ | 1 |
| $x_2$ | 2 | | $z_2$ | 2 | 1 | | $y_2$ | 2 |
| | | | $z_3$ | 2 | 2 | | | |

Thus, relation $S$ has three tuples $(1,1)$, $(2,1)$, $(2,2)$, and both $R$ and $T$ have two tuples $(1)$ and $(2)$. Each tuple is annotated with a Boolean variable $x_1, x_2, z_1, \dots, y_2$, which represents the independent event that the corresponding tuple is present in the database. The *lineage expression* $\varphi$ is then the DNF that states which tuples need to be present in order for the Boolean query $Q$ to be true:

$$\varphi = x_1 z_1 y_1 \vee x_2 z_2 y_1 \vee x_2 z_3 y_2$$

Calculating $\mathbb{P}[\varphi]$ for a general database instance is #P-hard. However, if we treat *each* occurrence of a variable $x_i$ in $\varphi$ as different (in other words, we dissociate $\varphi$ eagerly on all tuples $x_i$ from table $R$), then we get a read-once expression

$$\varphi' = x_1 z_1 y_1 \vee x'_{2,1} z_2 y_1 \vee x'_{2,2} z_3 y_2$$
$$= (x'_1 z_1 \vee x'_{2,1} z_2) y_1 \vee x'_{2,2} z_3 y_2$$

```
select IOR(Q3.P) as P                        create view VR as
from                                         select R.A,
    (select T.B, T.P*Q2.P as P                  1-power(1-R.P,1e0/count(*)) as P
    from T,                                  from R, S, T
        (select Q1.B, IOR(Q1.P) as P         where R.A=S.A
        from (select S.A, S.B, S.P*R.P as P  and S.B=T.B
            from R, S                        group by R.A, R.P
            where R.A = S.A) as Q1
        group by Q1.B) as Q2
    where T.B = Q2.B) as Q3
```

<div align="center">(a) Query $P_R$</div>                         (b) View $V_R$ for lower bound with $P_R$

Fig. 5.  (a): SQL query corresponding to plan $P_R$ for deriving an upper bound for the hard probabilistic Boolean query $Q :- R(X), S(X, Y), T(Y)$. Table $R$ needs to be replaced with the view $V_R$ from (b) for deriving a lower bound. IOR is a user-defined aggregate explained in the text and stated in Appendix C.

Writing $p_i, q_i, r_i$ for the probabilities of variables $x_i, y_i, z_i$, respectively, we can calculate

$$\mathbb{P}[\varphi'] = \big(\big((p'_1 \cdot r_1) \otimes (p'_{2,1} \cdot r_2)\big) \cdot q_1\big) \otimes (p'_{2,2} \cdot r_3 \cdot q_2)$$

where "$\cdot$" stands for multiplication and "$\otimes$" for independent-or.[10]

We know from Theorem 4.8 (3) that $\mathbb{P}[\varphi']$ is an upper bound to $\mathbb{P}[\varphi]$ by assigning the original probabilities to the dissociated variables. Furthermore, we have shown in [Gatterbauer et al. 2010] that $\mathbb{P}[\varphi]$ can be calculated with a *probabilistic query plan*

$$P_R = \pi_\emptyset^p \bowtie_Y^p \big[\pi_Y^p \bowtie_X^p \big[R(X), S(X, Y)\big], T(Y)\big]$$

where the probabilistic join operator $\bowtie^p [\dots]$ (in prefix notation) and the probabilistic project operator with duplicate elimination $\pi^p$ compute the probability assuming that their inputs are independent [Fuhr and Rölleke 1997]. Thus, when the join operator joins two tuples with probabilities $p_1$ and $p_2$, respectively, the output has probability $p_1 p_2$. When the independent project operator eliminates $k$ duplicate records with probabilities $p_1, \dots, p_k$, respectively, the output has probability $1 - \bar{p}_1 \cdots \bar{p}_k$. This connection between read-once formulas and query plans was first observed by Olteanu and Huang [2008]. We write here $P_R$ to emphasize that this plan dissociates tuples in table $R$.[11] Figure 5a shows the corresponding SQL query assuming that each of the input tables has one additional attribute P for the probability of a tuple. The query deploys a user-defined aggregate (UDA) IOR that calculates the independent-or for the probabilities of the tuples grouped together, i.e. $\mathsf{IOR}(p_1, p_2, \dots, p_n) = 1 - \bar{p}_1 \bar{p}_2 \cdots \bar{p}_n$. Appendix C states the UDA definition for PostgreSQL.

We also know from Theorem 4.8 (4) that $\mathbb{P}[\varphi']$ is a lower bound to $\mathbb{P}[\varphi]$ by assigning *new probabilities* $1 - \sqrt[3]{1 - p_2}$ to $x'_{2,1}$ and $x'_{2,2}$ (or more generally, any probabilities $p'_{2,1}$ and $p'_{2,2}$ with $\bar{p}'_{2,1} \cdot \bar{p}'_{2,2} \geq \bar{p}_2$). Because of the connection between the read-once expression $\varphi'$ and the query plan $P_R$, we can calculate the lower bound *by using the same SQL query* from Fig. 5a after exchanging the table $R$ with a view $V_R$ (Fig. 5b); $V_R$ is basically a copy of $R$ that replaces the probability $p_i$ of a tuple $x_i$ appearing in $\varphi$ with $1 - \sqrt[d_i]{1 - p_i}$ where $d_i$ is the number of times that $x_i$ appears in the lineage of $Q$. The view joins tables $R$, $S$ and $T$, groups the original input tuples $x_i$ from $R$, and assigns each $x_i$ the new probability $1 - \sqrt[d_i]{1 - p_i}$ calculated as 1-power(1-T.P,1e0/count(*)).

---

[10]The *independent-or* combines two probabilities as if calculating the disjunction between two independent events. It is defined as $p_1 \otimes p_2 := 1 - \bar{p}_1 \bar{p}_2$.

[11]Note we use the notation $P_R$ for both the probabilistic query plan and the corresponding SQL query.

Alternatively to $\varphi'$, if we treat each occurrence of a variable $y_j$ in $\varphi$ as different (in other words, we dissociate $\varphi$ eagerly on all tuples $y_j$ from table $T$), then we get another read-once expression

$$\varphi'' = x_1 z_1 y'_{1,1} \vee x_2 z_2 y'_{1,2} \vee x_2 z_3 y_2$$
$$= x_1 z_1 y'_{1,1} \vee x_2 (z_2 y'_{1,2} \vee z_3 y_2)$$

$\mathbb{P}[\varphi'']$ is an upper bound to $\mathbb{P}[\varphi]$ by assigning the original probabilities to the dissociated variables. In turn, $\mathbb{P}[\varphi'']$ can be calculated with another probabilistic query plan that dissociates all tuples from table $T$ instead of $R$:

$$P_T = \pi^p_\emptyset \bowtie^p_X \big[ R(X), \pi^p_X \bowtie^p_Y \big[ S(X,Y), T(Y) \big] \big]$$

Similarly to before, $\mathbb{P}[\varphi'']$ is a lower bound to $\mathbb{P}[\varphi]$ by assigning new probabilities $1 - \sqrt[2]{1 - q_1}$ to $y'_{1,1}$ and $y'_{1,2}$. And we can calculate this lower bound with the same query $P_T$ after exchanging $T$ with a view $V_T$ that replaces the probability $q_j$ of a tuple $y_j$ with $1 - \sqrt[d_j]{1 - q_j}$ where $d_j$ is the number of times that $y_j$ appears in the lineage of $Q$.

Note that both query plans will calculate upper and lower bounds to query $Q$ over *any database instance $D$*. In fact, all possible query plans give upper bounds to the true query probability. And as we have illustrated here, by replacing the input tables with appropriate views, we can use the same query plans to derive lower bounds. We refer the reader to [Gatterbauer et al. 2010] where we develop the theory of the partial dissociation order among all possible query plans and give a sound and complete algorithm that returns a set of query plans which are guaranteed to give the tightest bounds possible in a query-centric way for any conjunctive query without self-joins. For our example hard query $Q$, plans $P_R$ and $P_T$ are the best possible plans. We further refer to [Gatterbauer and Suciu 2013] for more details and an extensive discussion on how to speed up the resulting multi-query evaluation.

Also note that these upper and lower bounds can be derived with the help of *any standard relational database*, even cloud-based databases which commonly do not allow users to define their own UDAs.[12] To our best knowledge, this is the first technique that can upper and lower bound hard probabilistic queries *without any modifications to the database engine nor performing any calculations outside the database*.

# 7. ILLUSTRATIONS OF OBLIVIOUS BOUNDS

In this section, we study the quality of oblivious bounds across varying scenarios: We study the bounds as function of correlation between non-dissociated variables (Sect. 7.1), compare dissociation-based with model-based approximations (Sect. 7.2), illustrate a fundamental asymmetry between optimal upper and lower bounds (Sect. 7.3), show that increasing the number of simultaneous dissociations does not necessarily worsen the bounds (Sect. 7.4), and apply our framework to approximate hard probabilistic queries over TPC-H data with a standard relational database management system (Sect. 7.5).

## 7.1. Oblivious Bounds as Function of Correlation between Variables

*Example* 7.1 (*Oblivious Bounds and Correlations*).   Here we dissociate the DNF $\varphi_d = xA \vee xB$ and the analogous CNF $\varphi_c = (x \vee A)(x \vee B)$ on $x$ and study the er-ror of optimal oblivious bounds as function of the correlation between $A$ and $B$.[13] Clearly, the bounds also depend on the probabilities of the variables $x$, $A$, and $B$. Let $p = \mathbb{P}[x]$ and assume $A$ and $B$ have the same probability $q = \mathbb{P}[A] = \mathbb{P}[B]$. We set $p' = \mathbb{P}[x'_1] = \mathbb{P}[x'_2]$ according to the optimal symmetric bounds from Corollary 4.9.

---

[12]The UDA IOR can be expressed with standard SQL aggregates, e.g., "IOR(Q3.P)" can be evaluated with "1-exp(sum(log(case Q3.P when 1 then 1E-307 else 1-Q3.P end)))" on Microsoft SQL Azure.

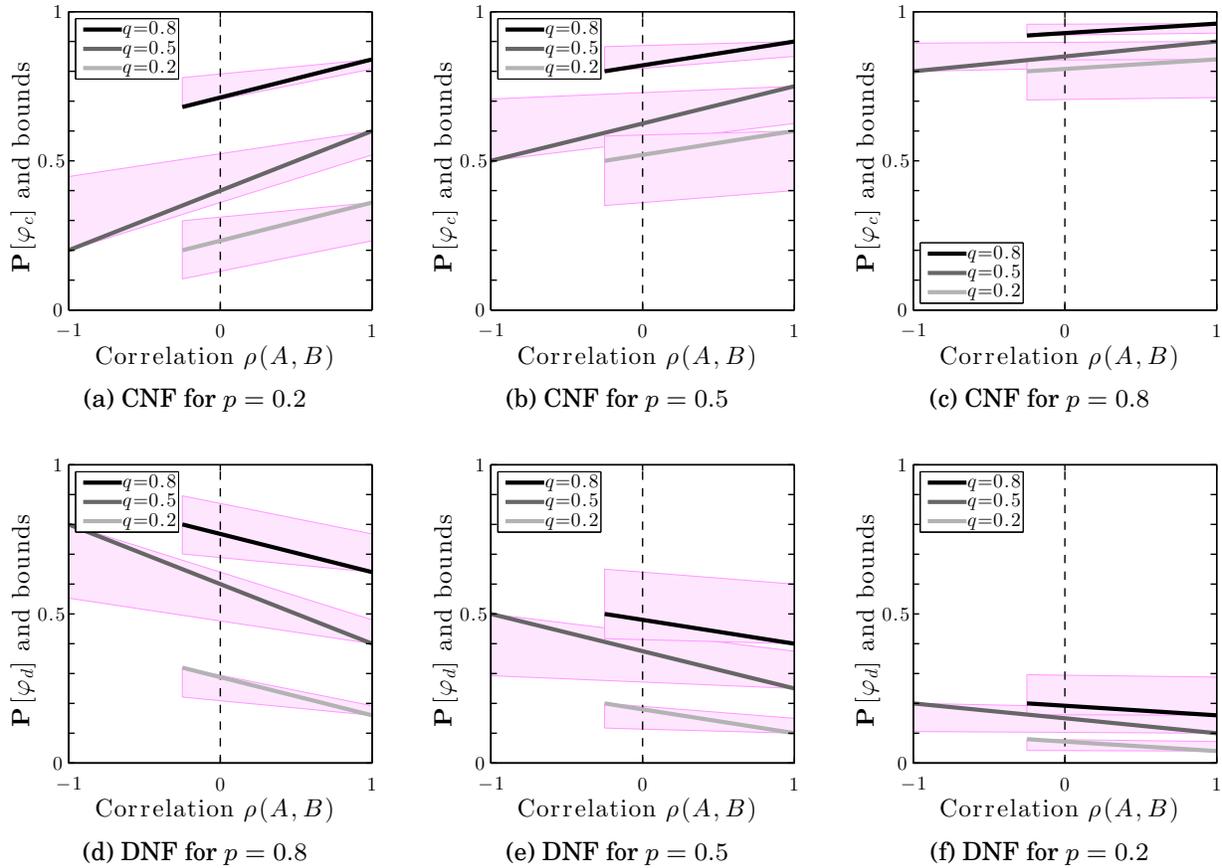

Fig. 6. Example 7.1. Probabilities of CNF $\varphi_c = (x \vee A)(x \vee B)$ and DNF $\varphi_d = xA \vee xB$ together with their symmetric optimal upper and lower oblivious bounds (borders of shaded areas) as function of the correlation $\rho(A, B)$ between $A$ and $B$, and parameters $p = \mathbb{P}[x]$ and $q = \mathbb{P}[A] = \mathbb{P}[B]$. For every choice of $p$, there are some $A$ and $B$ for which the upper or lower bound becomes tight.

In a few steps, one can calculate the probabilities as

$$\mathbb{P}[\varphi_d] = 2pq - p\mathbb{P}[AB]$$

$$\mathbb{P}[\varphi'_d] = 2p'q - p'^2\mathbb{P}[AB]$$

$$\mathbb{P}[\varphi_c] = p + (1-p)\mathbb{P}[AB]$$

$$\mathbb{P}[\varphi'_c] = 2p'q + p'^2(1-2q) + (1-p'^2)\mathbb{P}[AB]$$

*Results*: Figure 6 shows the probabilities of the expressions $\mathbb{P}[\varphi]$ (full lines) and those of their dissociations $\mathbb{P}[\varphi']$ (border of shaded areas) for various values of $p$, $q$ and as function of the correlation $\rho(A, B)$.[14] For example, Fig. 6d shows the case for DNF when $\mathbb{P}[x]$ is $p = 0.8$ and $A, B$ have the same probability $q$ of either 0.8, 0.5, or 0.2. When $A, B$

---

[13]Note that this simplified example also illustrates the more general case $\psi_d = xA \vee xB \vee C$ when $C$ is independent of $A$ and $B$, and thus $\mathbb{P}[\psi_d] = \mathbb{P}[\varphi_d](1 - \mathbb{P}[C]) + \mathbb{P}[C]$. As a consequence, the graphs in Fig. 6 for $\mathbb{P}[C] \neq 0$ would be vertically compressed and the bounds tighter in absolute terms.

[14]The *correlation* $\rho(A, B)$ between Boolean events $A$ and $B$ is defined as $\rho(A, B) = \frac{\text{cov}(A,B)}{\sqrt{\text{var}(A)\text{var}(B)}}$ with covariance $\text{cov}(A, B) = \mathbb{P}[AB] - \mathbb{P}[A]\mathbb{P}[B]$ and variance $\text{var}(A) = \mathbb{P}[A] - (\mathbb{P}[A])^2$ [Feller 1968]. Notice that

are not correlated at all ($\rho = 0$), then the upper bound is a better approximation when $q$ is small, and the lower bound is a better approximation when $q$ is large. On the other hand, if $A, B$ are not correlated, then there is no need to dissociate the two instances of $x$ as one can then compute $\mathbb{P}[(x \vee A)(x \vee B)]$ simply as $p + \bar{p}\,\mathbb{P}[A]\mathbb{P}[B]$. The more interesting case is when $A, B$ are positively correlated ($\mathbb{P}[AB] \geq \mathbb{P}[A]\mathbb{P}[B]$, e.g., positive Boolean functions of other independent variables $\mathbf{z}$, such as the provenance for probabilistic conjunctive queries). The right of the vertical dashed line of Fig. 6d shows that, in this case, dissociation offers very good upper and lower bounds, especially when the formula has a low probability. The graph also shows the effect of dissociation when $A, B$ are negatively correlated (left of dashed line). Notice that the correlation cannot always be $-1$ (e.g., two events, each with probability $> 0.5$, can never be disjunct). The graphs also illustrate why these bounds are obliviously optimal, i.e. without knowledge of $A, B$: for every choice of $p$, there are some $A, B$ for which the upper or lower bound becomes tight. ∎

## 7.2. Oblivious Bounds versus Model-based Approximations

*Example* 7.2 (*Disjunctive Dissociation and Models*).   This example compares the approximation of our dissociation-based approach with the model-based approach by Fink and Olteanu [2011] and illustrates how dissociation-based bounds are tighter, in general, than model-based approximations. For this purpose, we consider again the hard Boolean query $Q \coloneq R(X), S^d(X, Y), T(Y)$ over the database $D$ from Sect. 6. We now only assume that the table $S$ is deterministic, as indicated by the superscript $d$ in $S^d$. The query-equivalent lineage formula is then

$$\varphi = x_1 y_1 \vee x_2 y_1 \vee x_2 y_2$$

for which Fig. 7a shows the bipartite primal graph. We use this instance as its primal graph forms a $P_4$, which is the simplest 2-partite lineage that is not read-once.[15] In order to compare the approximation quality, we need to limit ourselves to an example which is tractable enough so we can generate the whole probability space. In practice, we allow each variable to have any of 11 discrete probabilities $D = \{0, 0.1, 0.2, \dots, 1\}$ and consider all $11^4 = 14641$ possible probability assignments $\nu : \langle p_1, p_2, q_1, q_2 \rangle \to D^4$ with $\mathbf{p} = \mathbb{P}[\mathbf{x}]$ and $\mathbf{q} = \mathbb{P}[\mathbf{y}]$. For each $\nu$, we calculate both the absolute error $\delta^* = \mathbb{P}[\varphi^*] - \mathbb{P}[\varphi]$ and the relative error $\varepsilon^* = \frac{\delta^*}{\mathbb{P}[\varphi]}$, where $\mathbb{P}[\varphi^*]$ stands for any of the approximations, and the exact probability $\mathbb{P}[\varphi]$ is calculated by the Shannon expansion on $y_1$ as $\varphi \equiv y_1(x_1 \vee x_2) \vee \neg y_1(x_2 y_2)$ and thus $\mathbb{P}_{\mathbf{p}, \mathbf{q}}[\varphi] = (1 - (1 - p_1)(1 - p_2))q_1 + (1 - q_1)p_2 q_2$.

*Models*: We use the model-based approach by Fink and Olteanu [2011] to approximate $\varphi$ with lowest upper bound (LUB) formulas $\varphi_{Ui}$ and greatest lower bound (GLB) formulas $\varphi_{Li}$, for which $\varphi_{Li} \Rightarrow \varphi$ and $\varphi \Rightarrow \varphi_{Ui}$, and neither of the upper (lower) bounds implies another upper (lower) bound. Among all models considered, we focus on only read-once formulas. Given the lineage $\varphi$, the 4 LUBs are $\varphi_{U1} = x_1 y_1 \vee x_2$, $\varphi_{U2} = y_1 \vee x_2 y_2$, $\varphi_{U3} = (x_1 \vee x_2)y_1 \vee y_2$, and $\varphi_{U4} = x_1 \vee x_2(y_1 \vee y_2)$. The 3 GLBs are $\varphi_{L1} = (x_1 \vee x_2)y_1$, $\varphi_{L2} = x_1(y_1 \vee y_2)$, and $\varphi_{L3} = x_1 y_1 \vee x_2 y_2$. For each $\nu$, we choose $\min_i(\mathbb{P}[\varphi_{Ui}])$ and $\max_i(\mathbb{P}[\varphi_{Li}])$ as the best upper and lower model-based bounds, respectively.

---

$\rho(A, B) = \frac{\mathbb{P}[AB] - q^2}{q - q^2}$ and, hence: $\mathbb{P}[AB] = \rho(A, B) \cdot (q - q^2) + q^2$. Further, $\mathbb{P}[AB] = 0$ (i.e. disjointness between $A$ and $B$) is not possible for $q > 0.5$, and from $\mathbb{P}[A \vee B] \leq 1$, one can derive $\mathbb{P}[AB] \geq 2q - 1$. In turn, $\rho = -1$ is not possible for $q < 0.5$, and it must hold $\mathbb{P}[AB] \geq 0$. From both together, one can derive the condition $\rho_{\min}(q) = \max(-\frac{q}{1-q}, -\frac{1 + q^2 - 2q}{q - q^2})$ which gives the minimum possible value for $\rho$, and which marks the left starting point of the graphs in Fig. 6 as function of $q$.

[15] A path $P_n$ is a graph with vertices $\{x_1, \dots, x_n\}$ and edges $\{x_1 x_2, x_2 x_3, \dots, x_{n1} x_n\}$.

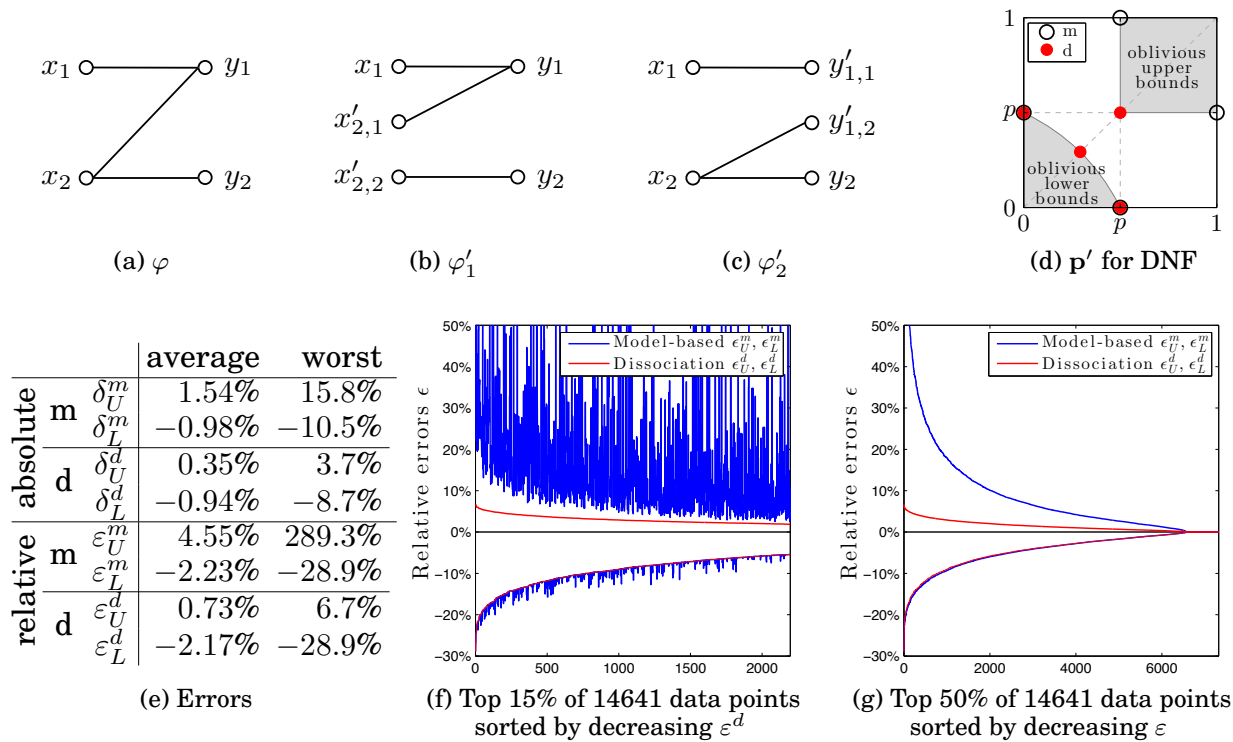

Fig. 7. Example 7.2. (a)-(c): Bipartite primal graphs for DNF $\varphi$ and two dissociations. Notice that the primal graphs of $\varphi'_1$ and $\varphi'_2$ are forests and thus correspond to read-once expressions. (d): For a given disjunctive dissociation d, there is only one optimal oblivious upper bound but infinitely many optimal lower bounds. We evaluate $\mathbb{P}[\varphi']$ for three of the latter (two of which coincide with models m) and keep the maximum as the best oblivious lower bound. (e): In comparison, dissociation gives substantially better *upper* bounds than model-based bounds (0.73% vs. 4.55% and 6.7% vs. 289.3% worst-case relative errors), yet *lower* bounds are only slightly better. (f): Relative errors of 4 approximations for individual data points sorted by the dissociation error for upper bounds and for lower bounds separately; this is why the dissociation errors show up as smooth curves (red) while the model based errors are unsorted and thus ragged (blue). (g): Here we sorted the errors for each approximation individually; this is why all curves are smooth.

*Dissociation:* Analogously, we consider the possible dissociations into read-once formulas. For our given $\varphi$, those are $\varphi'_1 = (x_1 \vee x'_{2,1})y_1 \vee x'_{2,2}y_2$ and $\varphi'_2 = x_1 y'_{1,1} \vee x_2(y'_{1,2} \vee y_2)$, with Fig. 7b and Fig. 7c illustrating the dissociated read-once primal graphs.[16] From Corollary 4.9, we know that $\mathbb{P}_{\mathbf{p}',\mathbf{q}}[\varphi'_1] \geq \mathbb{P}_{\mathbf{p},\mathbf{q}}[\varphi]$ for the only optimal oblivious upper bounds $p'_{2,1} = p'_{2,2} = p_2$ and $\mathbb{P}_{\mathbf{p}',\mathbf{q}}[\varphi'_1] \leq \mathbb{P}_{\mathbf{p},\mathbf{q}}[\varphi]$ for any $\mathbf{p}'_2$ with $\bar{p}'_{2,1}\bar{p}'_{2,2} = \bar{p}_2$. In particular, we choose 3 alternative optimal oblivious lower bounds $\mathbf{p}'_2 \in \{\langle p_2, 0\rangle, \langle 1 - \sqrt{1-p_2}, 1 - \sqrt{1-p_2}\rangle, \langle 0, p_2\rangle\}$ (see Fig. 7d). Analogously $\mathbb{P}_{\mathbf{p},\mathbf{q}'}[\varphi'_2] \geq \mathbb{P}_{\mathbf{p},\mathbf{q}}[\varphi]$ for $q'_{1,1} = q'_{1,2} = q_1$ and $\mathbb{P}_{\mathbf{p},\mathbf{q}'}[\varphi'_2] \leq \mathbb{P}_{\mathbf{p},\mathbf{q}}[\varphi]$ for $\mathbf{q}'_1 \in \{\langle q_1, 0\rangle, \langle 1-\sqrt{1-q_1}, 1-\sqrt{1-q_1}\rangle, \langle 0, q_1\rangle\}$. For each $\nu$, we choose the minimum among the 2 upper bounds and the maximum among the 6 lower bounds as the best upper and lower dissociation-based bounds, respectively.

*Results:* Figures 7e-g show that dissociation-based bounds are always better or equal to model-based bounds. The reason is that all model-based bounds are a special case of oblivious dissociation bounds. Furthermore, dissociation gives far better upper bounds, but only slighter better lower bounds. The reason is illustrated in Fig. 7d: the single dissociation-based upper bound $\mathbf{p}' = \langle p, p \rangle$ always dominates the two model-based upper bounds, whereas the two model-based lower bounds are special cases of infinitely many optimal oblivious lower dissociation-based bounds. As extreme case, it is therefore possible for a model-based lower bound to coincide with the best among all optimal

---

[16]Note that we consider here dissociation on both $x$- and $y$-variables, thus do not treat them as distinct.

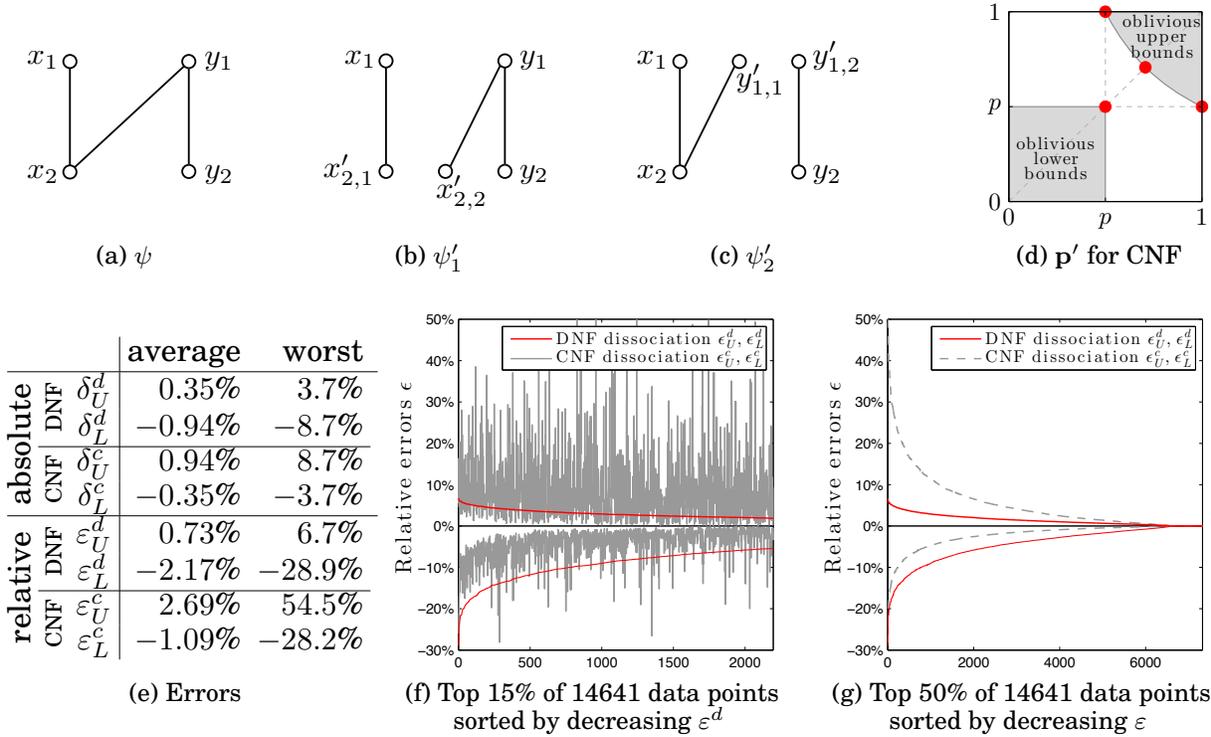

(a) $\psi$  (b) $\psi'_1$  (c) $\psi'_2$  (d) $\mathbf{p}'$ for CNF

| | | average | worst |
|---|---|---|---|
| absolute | $\delta^d_U$ DNF | 0.35% | 3.7% |
| | $\delta^d_L$ DNF | −0.94% | −8.7% |
| | $\delta^c_U$ CNF | 0.94% | 8.7% |
| | $\delta^c_L$ CNF | −0.35% | −3.7% |
| relative | $\varepsilon^d_U$ DNF | 0.73% | 6.7% |
| | $\varepsilon^d_L$ DNF | −2.17% | −28.9% |
| | $\varepsilon^c_U$ CNF | 2.69% | 54.5% |
| | $\varepsilon^c_L$ CNF | −1.09% | −28.2% |

(e) Errors

(f) Top 15% of 14641 data points sorted by decreasing $\varepsilon^d$

(g) Top 50% of 14641 data points sorted by decreasing $\varepsilon$

Fig. 8. Example 7.3. (a)-(c): Bipartite primal graphs for CNF $\psi$ and two dissociations. (d): For a given conjunctive dissociation c, we use the only optimal oblivious lower bound and three of infinitely many optimal oblivious upper bounds. (e): In comparison, disjunctive dissociation gives better *upper* bounds than conjunctive dissociation (0.73% vs. 2.69% average and 6.7% vs. 54.5% worst-case relative errors), and v.v. for lower bounds. (f): Relative errors of 4 approximations for individual data points sorted by the disjunctive dissociation error $\varepsilon^d$ for upper bounds and for lower bounds separately; this is why the DNF dissociation errors show up as smooth curves (red) while the CNF dissociation errors are unsorted and thus ragged (gray). (g): Here we sorted the errors for each approximation individually; this is why all curves are smooth.

oblivious lower dissociation bounds. For our example, we evaluate three oblivious lower bounds, two of which coincide with models. ∎

## 7.3. Conjunctive versus Disjunctive Dissociation

*Example* 7.3 (*Disjunctive and Conjunctive Dissociation*). This example illustrates an interesting asymmetry: optimal upper bounds for disjunctive dissociations and optimal lower bounds for conjunctive dissociations are not only unique but also better, on average, than optimal upper bounds for conjunctive dissociations and optimal lower bounds for disjunctive dissociations, respectively (see Figure 1). We show this by comparing the approximation of a function by either dissociating a conjunctive or a disjunctive expression for the same function.

We re-use the setup from Example 7.2 where we had a function expressed by a disjunctive expression $\varphi$. Our DNF $\varphi$ can be written as CNF $\psi = (x_1 \vee x_2)(y_1 \vee x_2)(y_1 \vee y_2)$ with $f_\varphi = f_\psi$[17], and two conjunctive dissociations $\psi'_1 = (x_1 \vee x'_{2,1})(y_1 \vee x'_{2,2})(y_1 \vee y_2)$ and $\psi'_2 = (x_1 \vee x_2)(y'_{1,1} \vee x_2)(y'_{1,2} \vee y_2)$ (Figures 8a-c shows the primal graphs). Again from Corollary 4.9, we know that $\mathbb{P}_{\mathbf{p}', \mathbf{q}}[\varphi'_1] \leq \mathbb{P}_{\mathbf{p}, \mathbf{q}}[\varphi]$ for the only optimal oblivious lower bounds $p'_{2,1} = p'_{2,2} = p_2$ and $\mathbb{P}_{\mathbf{p}', \mathbf{q}}[\varphi'_1] \geq \mathbb{P}_{\mathbf{p}, \mathbf{q}}[\varphi]$ for any $\mathbf{p}'_2$ with $p'_{2,1}p'_{2,2} = p_2$. In particular, we choose 3 alternative optimal oblivious lower bounds $\mathbf{p}'_2 \in \{\langle p_2, 1 \rangle, \langle \sqrt{p_2}, \sqrt{p_2} \rangle, \langle 1, p_2 \rangle\}$ (see Fig. 7d). Analogously $\mathbb{P}_{\mathbf{p}, \mathbf{q}'}[\varphi'_2] \leq \mathbb{P}_{\mathbf{p}, \mathbf{q}}[\varphi]$ for $q'_{1,1} = q'_{1,2} = q_1$ and $\mathbb{P}_{\mathbf{p},\mathbf{q}'}[\varphi'_2] \geq \mathbb{P}_{\mathbf{p},\mathbf{q}}[\varphi]$ for $\mathbf{q}'_1 \in \{\langle q_1, 1\rangle, \langle\sqrt{q_1}, \sqrt{q_1}\rangle, \langle 1, q_1\rangle\}$. For each $\nu$, we choose the maximum among the 2 lower bounds and the minimum among the 6 upper bounds as the best upper and lower conjunctive dissociation-based bounds, respectively. We then compare with the approximations from the DNF $\varphi$ in Example 7.2.

---

[17]Notice that this transformation from DNF to CNF is hard, in general. We not do not focus on algorithmic aspects in this paper, but rather show the potential of this new approach.

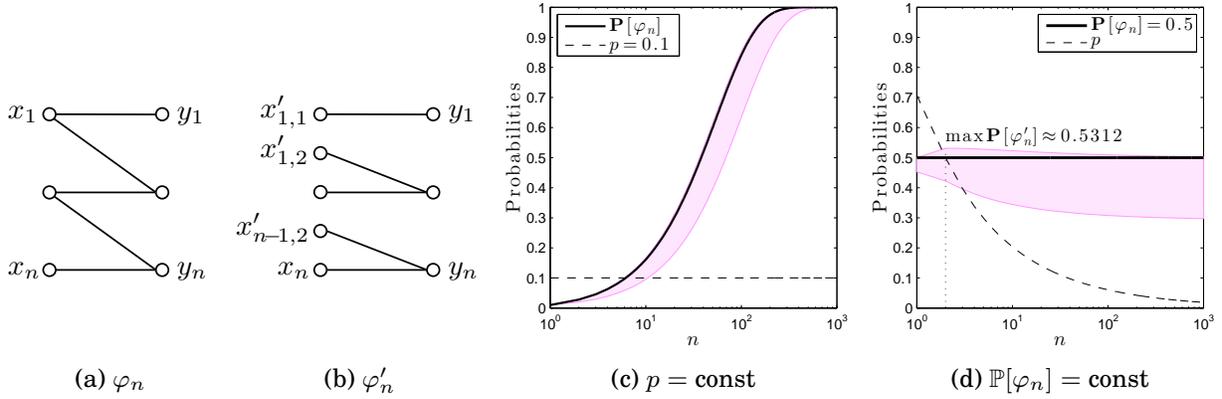

Fig. 9. Example 7.4. (a), (b): Primal graphs for path $P_n$ DNF $\varphi_n$ and its dissociation $\varphi'_n$. (c), (d): $\mathbb{P}[\varphi_n]$ together with their symmetric optimal upper and lower oblivious bounds (borders of shaded areas) as function of $n$. (c) keeps $p = 0.1$ constant, whereas (d) varies $p$ so as to keep $\mathbb{P}[\varphi] = 0.5$ constant for increasing $n$. The upper bounds approximate the probability of the DNF very well and even become tight for $n \to \infty$.

*Results:* Figures 8e-g show that optimal disjunctive upper bounds are, in general but not consistently, better than optimal conjunctive upper bounds for the same function ($\approx 83.5\%$ of those data points with different approximations are better for conjunctive dissociations). The dual result holds for lower bounds. This duality can be best seen in the correspondences of absolute errors between upper and lower bounds. ∎

## 7.4. Multiple Dissociations at Once

Here we investigate the influence of the primal graph and number of dissociations on the tightness of the bounds. Both examples correspond to the lineage of the standard unsafe query $Q :- R(X), S(X, Y), T(Y)$ over two different database instances.

*Example* 7.4 (*Path $P_n$ as Primal Graph*). This example considers a DNF expression whose primal graph forms a $P_n$, i.e. a path of length $n$ (see Fig. 9a). Note that this is a generalization of the path $P_4$ from Example 7.2 and corresponds to the lineage of the same unsafe query over larger database instance with $2n - 1$ tuples:

$$\varphi_n = x_1 y_1 \vee x_1 y_2 \vee x_2 y_2 \vee \ldots \vee x_{n-1} y_n \vee x_n y_n$$
$$\varphi'_n = x'_{1,1} y_1 \vee x'_{1,2} y_2 \vee x'_{2,1} y_2 \vee \ldots \vee x'_{n-1,2} y_n \vee x'_n y_n$$

*Exact*: In the following, we assume the probabilities of all variables to be $p$ and use the notation $p_n := \mathbb{P}[\varphi_n]$ and $p^*_n := \mathbb{P}[\varphi^*_n]$, where $\varphi^*_n$ corresponds to the formula $\varphi_n$ without the last conjunct $x_n y_n$. We can then express $p_n$ as function of $p^*_n$, $p_{n-1}$ and $p^*_{n-1}$ by recursive application of Shannon's expansion to $x_n$ and $y_n$:

$$p_n = \mathbb{P}[x_n]\big(\mathbb{P}[y_n] + \mathbb{P}[\bar{y}_n]p_{n-1}\big) + \mathbb{P}[\bar{x}_n]p^*_n$$
$$p^*_n = \mathbb{P}[y_n]\big(\mathbb{P}[x_{n-1}] + \mathbb{P}[\bar{x}_{n-1}]p^*_{n-1}\big) + \mathbb{P}[\bar{y}_n]p_{n-1}$$

We thus get the linear recurrence system

$$p_n = A_1 p_{n-1} + B_1 p^*_{n-1} + C_1$$
$$p^*_n = A_2 p_{n-1} + B_2 p^*_{n-1} + C_2$$

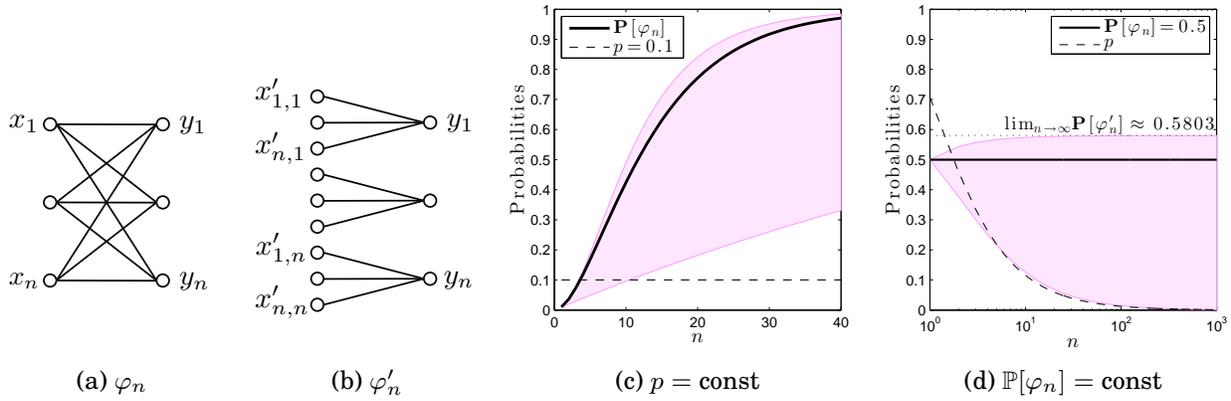

**Fig. 10.** Example 7.5. (a), (b): Primal graphs for complete bipartite DNF $\varphi_n$ and its dissociation $\varphi_n'$. (c), (d): $\mathbb{P}[\varphi_n]$ together with their symmetric optimal upper and lower oblivious bounds (borders of shaded areas) as function of $n$. (d) varies $p$ so as to keep $\mathbb{P}[\varphi] = 0.5$ constant for increasing size $n$. The oblivious upper bound is ultimately bounded despite having $n^2$ fresh variables in the dissociated DNF for increasing $n$.

with $A_1 = \bar{p}$, $B_1 = p\bar{p}^2$, $C_1 = p^2(2-p)$, $A_2 = \bar{p}$, $B_2 = p\bar{p}$, and $C_2 = p^2$. With a few manipulations, this recurrence system can be transformed into a linear non-homogenous recurrence relation of second order $p_n = Ap_{n-1} + Bp_{n-2} + C$ where $A = A_1 + B_2 = \bar{p}(1+p)$, $B = A_2B_1 - A_1B_2 = -p^2\bar{p}^2$, and $C = B_1C_2 + C_1(1 - B_2) = p^2(p\bar{p}^2 + (2-p)(1-p\bar{p}))$. Thus we can recursively calculate $\mathbb{P}[\varphi_n]$ for any probability assignment $p$ starting with initial values $p_1 = p^2$ and $p_2 = 3p^2 - 2p^3$.

*Dissociation*: Figure 9b shows the primal graph for the dissociation $\varphi_n'$. Variables $x_1$ to $x_{n-1}$ are dissociated into two variables with same probability $p'$, whereas $x_n$ into one with original probability $p$. In other words, with increasing $n$, there are more variables dissociated into two fresh ones each. The probability $\mathbb{P}[\varphi_n']$ is then equal to the probability that at least one variable $x_i$ is connected to one variable $y_j$:

$$\mathbb{P}[\varphi_n'] = 1 - \left(1 - pp'\right)\left(1 - p(1 - \bar{p}'^2)\right)^{n-2}\left(1 - p(1 - \bar{p}\bar{p}')\right)$$

We set $p' = p$ for upper bounds, and $p' = 1 - \sqrt{1-p}$ for lower bounds.

*Results*: Figures 9c-d shows the interesting result that the disjunctive upper bounds become tight for increasing size of the primal graph, and thus increasing number of dissociations. This can be best seen in Fig. 9d for which $p$ is chosen as to keep $\mathbb{P}[\varphi] = 0.5$ constant for varying $n$ and we have $\lim_{n\to\infty} \mathbb{P}[\varphi_n'] = \mathbb{P}[\varphi_n] = 0.5$ for upper bounds. In contrast, the disjunctive lower bounds become weaker but still have a limit value $\lim_{n\to\infty} \mathbb{P}[\varphi_n'] \approx 0.2929$ (derived numerically). ∎

*Example* 7.5 (*Complete bipartite graph $K_{n,n}$ as Primal Graph*). This example considers a DNF whose primal graph forms a complete bipartite graph of size $n$, i.e. each variable $x_i$ is appearing in one clause with each variable $y_j$ (see Fig. 10a). Note that this example corresponds to lineage for the standard unsafe query over a database instance with $\mathcal{O}(n^2)$ tuples:

$$\varphi_n = \bigvee_{(i,j)\,\in\,[n]^2} x_iy_j$$

$$\varphi_n' = \bigvee_{j\in[n]} \left(y_j \bigvee_{i\in[n]} x_{i,j}'\right)$$

*Exact*: We again assume that all variables have the same probability $p = \mathbb{P}[x_i] = \mathbb{P}[y_i]$. $\mathbb{P}[\varphi_n]$ is then equal to the probability that there is at least one tuple $x_i$ and at

```
select distinct s_nationkey           Supplier(s_suppkey, s_nationkey)
from Supplier, Partsupp, Part         PartSupp(ps_suppkey, ps_partkey)
where s_suppkey = ps_suppkey          Part(p_partkey, p_name)
and ps_partkey = p_partkey
and s_suppkey <= $1
and p_name like $2
```

(a) Deterministic query $Q(a)$                    (b) Relevant TPC-H schema

Fig. 11. Example 7.6. Parameterized SQL query $Q(a)$ and relevant portion of the TPC-H schema.

least one tuple $y_i$:

$$\mathbb{P}[\varphi_n] = \big(1 - (1 - p)^n\big)^2 \tag{6}$$

*Dissociation*: Figure 10b shows the primal graph for the dissociation $\varphi'_n$. Each variable $x_i$ is dissociated into $n$ fresh variables with same probability $p'$, i.e. there are $n^2$ fresh variables in total. The probability $\mathbb{P}[\varphi'_n]$ is then equal to the probability that at least one variable $y_i$ is connected to one variables $x_{i,j}$:

$$\mathbb{P}[\varphi'_n] = 1 - \Big(1 - p\big(1 - (1 - p')^n\big)\Big)^n$$

We will again choose $p$ as to keep $r := \mathbb{P}[\varphi_n]$ constant with increasing $n$, and then calculate $\mathbb{P}[\varphi'_n]$ as function of $r$. From Eq. 6, we get $p = 1 - \sqrt[n]{1 - \sqrt{r}}$ and then set $p' = p$ for upper bounds, and $p' = 1 - \sqrt[n]{1 - p}$ for lower bounds as each dissociated variable is replaced by $n$ fresh variables. It can then be shown that $\mathbb{P}[\varphi'_n]$ for the upper bound is monotonically increasing for $n$ and bounded below 1 with the limit value:

$$\lim_{n \to \infty} \mathbb{P}[\varphi'_n] = 1 - (1 - \sqrt{r})^{\sqrt{r}}$$

*Results*: Figure 10d keeps $\mathbb{P}[\varphi] = 0.5$ constant (by decreasing $p$ for increasing $n$) and shows the interesting result that the optimal upper bound is itself upper bounded and reaches a limit value, although there are more variables dissociated, and each variable is dissociated into more fresh ones. This limit value is $0.5803$ for $r = 0.5$. However, lower bounds are not useful in this case. ■

## 7.5. Dissociation with a Standard Relational Database Management System

*Example* 7.6 (*TPC-H*). Here we apply the theory of dissociation to bound hard probabilistic queries with the help of PostgreSQL 9.2, an open-source relational DMBS.[18] We use the TPC-H DBGEN data generator[19] to generate a 1GB database. We then add a column P to each table, and assign to each tuple either the probability $p = 0.1$, or $p = 0.5$, or a random probability from the set $\{0.01, 0.02, \dots, 0.5\}$ ("$p =$ rand 0.5"). Choosing tuple probabilities $p \leq 0.5$ helps us avoid floating-point errors for queries with very large lineages whose query answer probabilities would otherwise be too close to 1.[20] Our experiments use the following parameterized query (Fig. 11):

$$Q(a) :\!- S(\underline{s}, a), PS(s, u), P(\underline{u}, n), s \leq \$1, n \text{ like } \$2$$

Relations $S$, $PS$ and $P$ represent tables Supplier, PartSupp and Part, respectively. Variable $a$ stands for attribute nationkey ("answer tuple"), $s$ for suppkey, $u$ for partkey ("unit"),

---

```
create view VP as                          select s_nationkey, IOR(Q3.P) as P
select p_partkey, s_nationkey,             from
    1-power(1-P.P,1e0/count(*)) as P          (select S.s_nationkey, S.P*Q2.P as P
from Part P, Partsupp, Supplier               from Supplier S,
where p_partkey=ps_partkey                       (select Q1.ps_suppkey, s_nationkey, IOR(Q1.P) as P
and ps_suppkey = s_suppkey                       from
and s_suppkey <= $1                                 (select ps_suppkey, s_nationkey, PS.P*VP.P as P
and p_name like $2                                  from Partsupp PS, VP
group by p_partkey, s_nationkey, P.P                where ps_partkey = p_partkey
                                                    and ps_suppkey <= $1) as Q1
                                              group by Q1.ps_suppkey, s_nationkey) as Q2
                                           where s_suppkey = Q2.ps_suppkey) as Q3
                                           group by Q3.s_nationkey
```

(a) View $V_P$ for lower bounds with $P_P^*$                    (b) Query $P_P^*$

Fig. 12. Example 7.6. (a) View definition $V_P$ and (b) and query $P_P^*$ for deriving the lower bounds by dissociating table $P$. Note the inclusion of the attribute nationkey in $V_P$ for reasons explained in the text.

and $n$ for name. The probabilistic version of this query asks which nations (as determined by the attribute nationkey) are most likely to have suppliers with suppkey $\leq$ \$1 that supply parts with a name like \$2 when all tuples in Supplier, PartSupp, and Part are probabilistic. Parameters \$1 and \$2 allow us to reduce the number of tuples that can participate from tables Supplier and Part, respectively, and to thus study the effects of the lineage size on the predicted dissociation bounds and running times. By default, tables Supplier, Partsupp and Part have 10k, 800k and 200k tuples, respectively, and there are 25 different numeric attributes for nationkey. For parameter \$1, we choose a value $\in \{500, 1000, \ldots, 10000\}$ which selects the respective number of tuples from table Supplier. For parameter \$2, we choose a value $\in \{`\%`, `\%red\%`, `\%red\%green\%`\}$ which selects 200k, 11k or 251 tuples in table Part, respectively.

*Translation into SQL*: Note that the lineage for each individual query answer corresponds to the lineage for the Boolean query $Q$ from Sect. 6, which we know is hard, in general. We thus bound the probabilities of the query answers by evaluating *four different queries* that correspond to the query-centric dissociation bounds from Sect. 6: dissociating either table Supplier or table Part, and calculating either upper and lower bounds. To get the final upper (lower) bounds, we take the minimum (maximum) of the two upper (lower) bounds for each answer tuple. The two query plans are as follows:

$$P_S(a) = \pi_a^p \bowtie_u^p \big[\pi_{a,u}^p \bowtie_s^p [S(\underline{s}, a), PS(s, u), s \leq \$1], P(\underline{u}, n), n \text{ like } \$2\big]$$

$$P_P(a) = \pi_a^p \bowtie_s^p \big[S(\underline{s}, a), \pi_s^p \bowtie_u^p \big[PS(s, u), s \leq \$1, P(\underline{u}, n), n \text{ like } \$2\big]\big]$$

Notice one technical detail for determining the lower bounds with plan $P_P$: Any tuple $t$ from table Part may appear a different number of times in the lineages for different query answers.[21] Thus, for every answer $a$ that has $t$ in its lineage, we need to create a *distinct copy of* $t$ in the view $V_P$, with a probability that depends only on the number of times that $t$ appears in the lineage for $a$.[22] Thus, the view definition for $V_R$ needs to include the attribute nationkey (Fig. 12a) and $P_P$ needs to be adapted as follows (Fig. 12b):

$$P_P^*(a) = \pi_a^p \bowtie_s^p \big[S(\underline{s}, a), \pi_{s,a}^p \bowtie_u^p \big[PS(s, u), s \leq \$1, VP(\underline{u}, a)\big]\big]$$

---

[21]Lower bounds by dissociating Supplier are easier since the table includes the query answer attribute nationkey. As consequence, any tuple from Supplier may appear in the lineage of one query answer only.
[22]We could actually use the total number of times the tuple appears in all lineages and still get lower bounds. However, the resulting lower bounds would be weaker and not obliviously optimal.

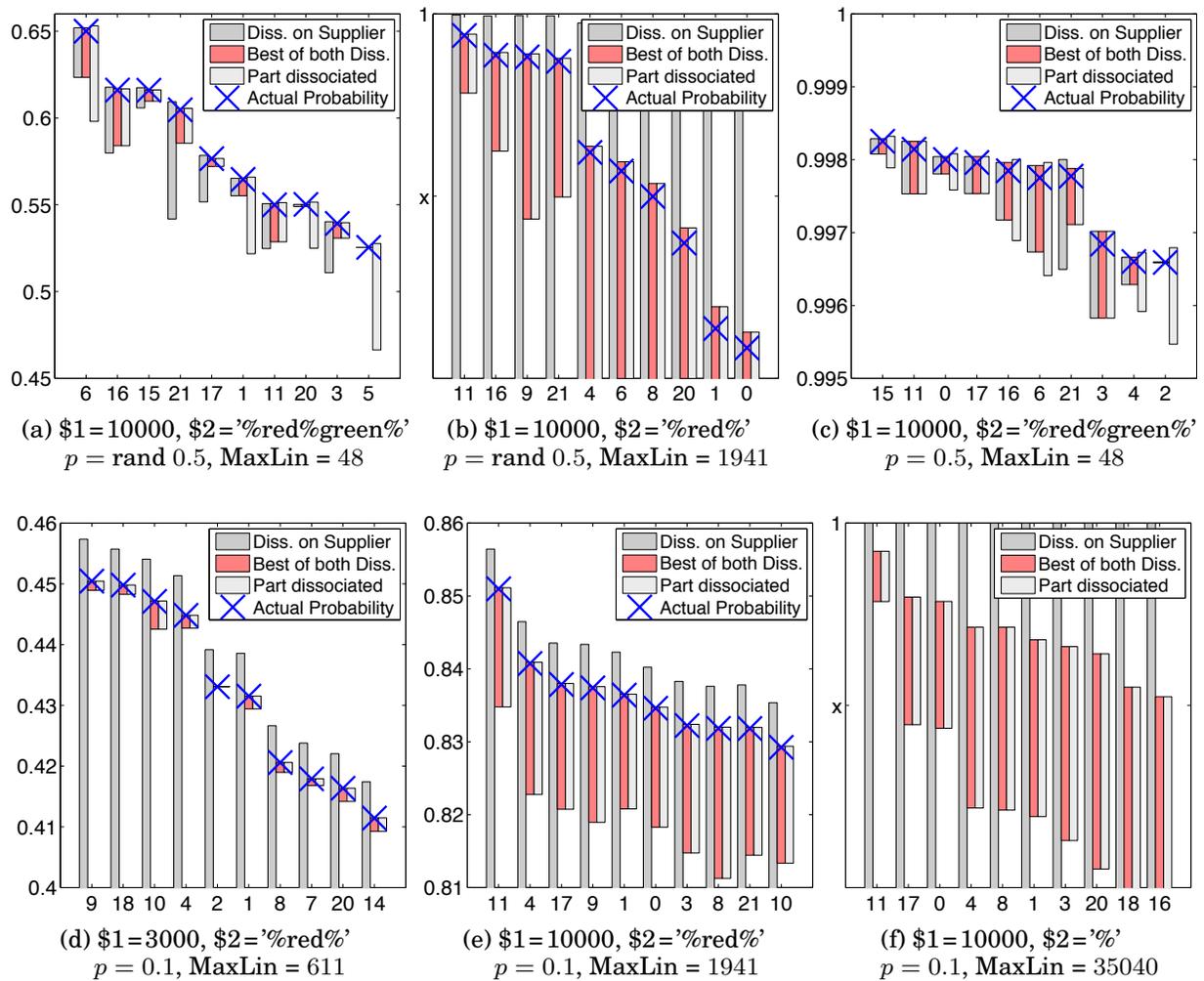

(a) $1 = 10000, $2 = '%red%green%'
$p$ = rand 0.5, MaxLin = 48

(b) $1 = 10000, $2 = '%red%'
$p$ = rand 0.5, MaxLin = 1941

(c) $1 = 10000, $2 = '%red%green%'
$p = 0.5$, MaxLin = 48

(d) $1 = 3000, $2 = '%red%'
$p = 0.1$, MaxLin = 611

(e) $1 = 10000, $2 = '%red%'
$p = 0.1$, MaxLin = 1941

(f) $1 = 10000, $2 = '%'
$p = 0.1$, MaxLin = 35040

| Case | Answer | Fig. | #Part | %diss. | #fresh | #Supp. | %diss. | #fresh | tighter bounds |
|------|--------|------|-------|--------|--------|--------|--------|--------|----------------|
| (A)  | (2)    | (c)  | 40    | 7.5%   | 2      | 43     | 0.0%   | -      | $P_S$          |
| (B)  | (6)    | (a)  | 42    | 11.9%  | 2      | 53     | 7.5%   | 2      | $P_S$          |
| (C)  | (11)   | (b)  | 1830  | 5.8%   | 2      | 434    | 95.6%  | 2-11   | $P_P$          |
| (D)  | (11)   | (f)  | 32899 | 6.3%   | 2      | 438    | 100.0% | 80     | $P_P$          |

(g) Overview of 4 cases discussed in the text

Fig. 13. Example 7.6. Probabilities for the top 10 query answers for varying query parameters $1, $2, and tuple probabilities $p$. The ranking is determined by the upper dissociation bounds (upper end of the red interval) and is identical to the one determined by the actual probabilities (crosses), except in (c) where tuples 6 and 21 are flipped. MaxLin shows the maximal lineage among the query answers, which is too big in (f) for exact probabilistic inference. (b): $x = 0.999\,999\,999\,999 = 1 - 10^{-12}$. (f): $x = 0.999\,999\,999\,9 = 1 - 10^{-10}$.

In order to speed up the resulting multi-query evaluation, we apply a determinis­tic semi-join reduction on the input tables and then reuse intermediate query results across all four subsequent queries. Since query optimization is not the focus of this paper, we do not show these techniques in Fig. 12 and instead refer to [Gatterbauer and Suciu 2013] for a detailed discussion of techniques to speed up query dissociation. In addition, the exact SQL statements that allow the interested reader to repeat our experiments with PostgreSQL are available on the LaPushDB project page.[23]

---

[23] http://LaPushDB.com/

*Ground truth*: To compare our bounds against ground truth, we construct the lineage DNF for each query answer and use DeMorgan to write it as *"dual lineage CNF"* without exponential increase in size. For example, the DNF $\varphi = x_1 x_3 \vee \bar{x}_1 x_2$ can be written as CNF $\bar{\varphi} = (\bar{x}_1 \vee \bar{x}_3) \wedge (\bar{x}_1 \vee \bar{x}_2)$ with $\mathbb{P}[\bar{\varphi}] = 1 - \mathbb{P}[\varphi]$. The problem of evaluating a probabilistic CNF can further be translated into the problem of computing the partition function of a propositional Markov random field, for which the AI community has developed sophisticated solvers. For our experiments, we use a tool called SampleSearch [Gogate and Domingos 2010; Gogate and Dechter 2011][24].

*Quality Results*: Figure 13 shows the top 10 query answers, as predicted by the upper dissociation bounds for varying query parameters $1 and $2, as well as varying tuple probabilities $p$. The crosses show the ground truth probabilities as determined by SampleSearch. The red intervals shows the interval between upper and lower dissociation bounds. Recall that the final dissociation interval is the intersection between the interval from dissociation on Supplier (left of the red interval) and on Part (right of the red interval). The graphs suggest that the *upper dissociation bounds are very close to the actual probabilities*, which is reminiscent of Sect. 7.4 having shown that upper bounds for DNF dissociations are commonly closer to the true probabilities than lower bounds. For Fig. 13f, we have no ground truth as the lineage for the top tuple (11) has size 32899 (i.e. the corresponding DNF has 32899 clauses), which is too big for exact probabilistic inference. However, extrapolating from Fig. 13d and Fig. 13e to Fig. 13f, we speculate that upper dissociation bounds give good approximations here as well.

The different interval sizes (i.e. quality of the bounds) arise from different numbers of dissociated tuples in the respective lineages. We illustrate with 4 cases (Fig. 13g):

(A) If there is a plan that does not dissociate any tuple, then both upper and lower bounds coincide. This scenario is also called *data-safe* [Jha et al. 2010]. For example, the lineage for answer (2) in Fig. 13c has 40 unique tuples from table Part, out of which 3 (7.5%) are dissociated into 2 fresh ones with $P_P$. However, all of the 43 tuples from table Supplier that appear in the lineage appear only once. Therefore, $P_S$ dissociates *no tuple* when calculating the answer probability for (2), and as a result gives us the exact value.

(B) The lineage for the top-ranked answer (6) in Fig. 13a has 42 unique tuples from table Part, out of which 5 ($\approx 11.9\%$) are dissociated into 2 fresh ones with $P_P$. In contrast, the lineage has 53 unique tuples from table Supplier, out of which only 4 ($\approx 7.5\%$) are dissociated into 2 fresh ones with $P_S$. Intuitively, $P_S$ should give us tighter bounds, which is confirmed by the results.

(C) The lineage for the top-ranked answer (11) in Fig. 13b and Fig. 13e (both figures show results for the same query, but for different tuple probabilities $p = 0.1$ or $p = \text{rand } 0.5$) has 1830 unique tuples from table Part, out of which $\approx 5.8\%$ are dissociated into 2 or 3 fresh ones with $P_P$. In contrast, the lineage has only 434 unique tuples from table Supplier, out of which $\approx 95.6\%$ are dissociated into into 2 to 11 fresh variables. Thus, $P_P$ gives far tighter bounds in this case.

(D) The lineage for the same answer (11) in Fig. 13f has 32899 unique tuples from table Part, out of which $\approx 6.3\%$ are dissociated into 2-4 fresh ones with $P_P$. In contrast, the lineage has only 438 unique tuples from table Supplier, out of which *all* are dissociated into 80 fresh ones with $P_S$ (this is an artifact of the TPC-H random database generator). Thus, the bounds for $P_S$ are very poor.

Importantly, *relevance ranking of the answer tuples by upper dissociation bounds* approximates the ranking by query reliability very well. For example, for the case of $p = 0.1$ and $2='\%\text{red}\%'$ (Fig. 13d and Fig. 13e), the ranking given by the minimum upper bounds was identical to the ranking given by the ground truth for all parameter choices $\$1 \in \{500, 1000, \ldots, 10000\}$. Figure 13c shows an example of an incorrect rankings (for parameters $1=10000, $2='%red%green%', $p = 0.5$): Here tuples 6 and 21 are flipped as compared to their actual probabilities 0.99775 and 0.99777, respectively.

---

[24]http://www.hlt.utdallas.edu/~vgogate/SampleSearch.html

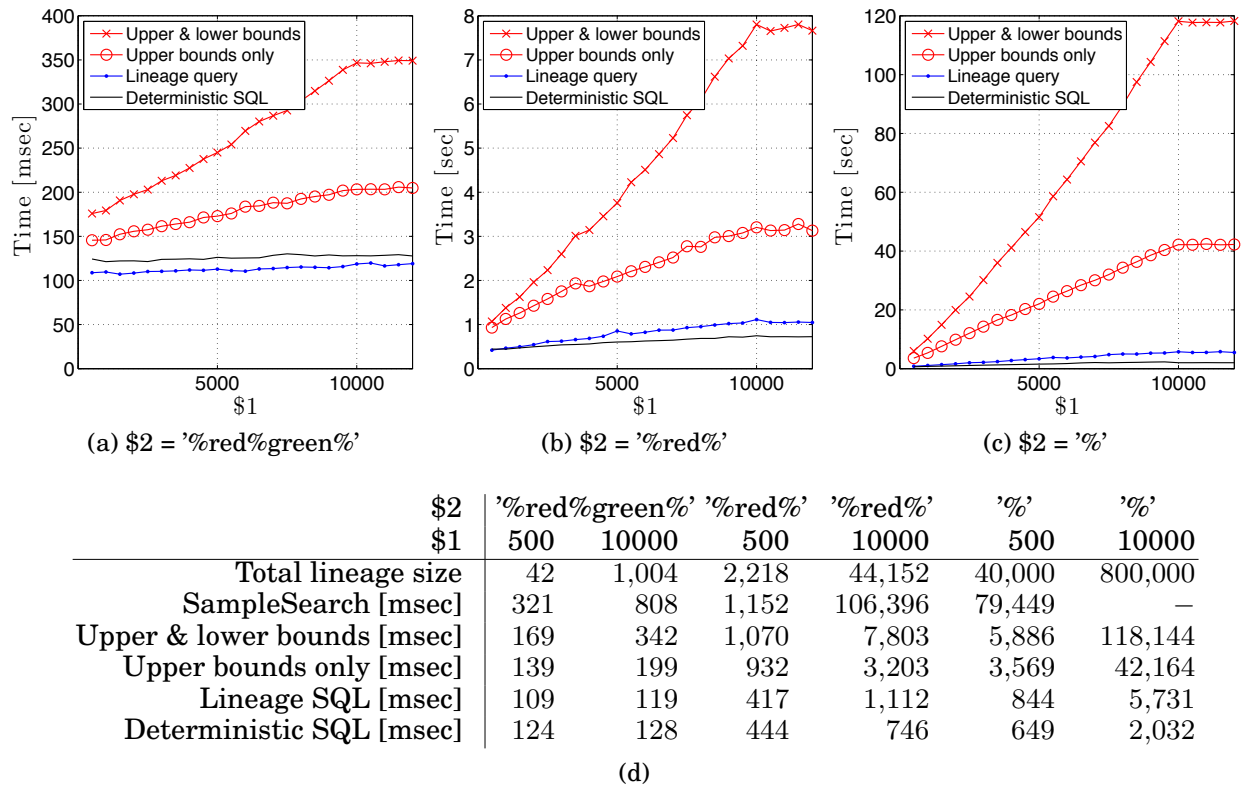

Fig. 14. Example 7.6. Timing results for queries with varying parameters $1 and $2: For small lineage ($< 10000$), dissociation bounds can be calculated in a small multiple ($< 4$) of the time needed for a standard deterministic query. For large lineages ($> 10000$), calculation scales linearly in the size of the total lineage.

*Timing Results*: Figure 14 compares the times for evaluating the deterministic query (Fig. 11a) with the times for calculating the dissociation bounds for changing parameters $1 and $2. As experimental platform, we use PostgreSQL 9.2 on a 2.5 Ghz Intel Core i5 with 16G of main memory. We run each query 5 times and take the average execution time. Figure 14d also shows the size of the total lineage of a query (which is the same as the number of query results for the deterministic query without projection) and the times needed by SampleSearch to evaluate the ground truth, if possible.[25] Since table Supplier contains exactly 10k tuples with suppkey $\in \{1, \ldots, 10000\}$, any choice of $\$1{\geq}10000$ has no effect on the query. We show separate graphs for the time needed to calculate the *upper bounds only* (which our theory and experiments suggest give better absolute approximations) and the time for both *upper and lower bounds* (lower bounds are more expensive due to the required manipulation of the input tuples). We also show the times for retrieving the lineage with a *lineage query*. Any probabilistic approach that evaluates the probabilities outside of the database engine needs to issue this query to retrieve the DNF for each answer. The time needed for the lineage query thus serves as minimum benchmark for *any* probabilistic approximation.

Our timing results show that, for small lineages ($< 10000$), calculating upper and lower bounds can be achieved in a time that is only a small multiple ($< 4$) of the time needed for an equivalent deterministic query. For large lineages ($> 10000$), calculating the bounds scales linearly with the size of the lineage (Fig. 14c), whereas determin-istic query evaluation can scale in sublinear time (recall that the cardinality of the answer set is maximal 25 across all queries since the database contains only 25 different values for the answer attribute nationkey). Importantly, scalability for dissociation is *independent of the tractability of the data instance*, e.g. maximum treewidth of the lineage for any query answer. In contrast, SampleSearch quickly takes too long for increasing lineage. For example, SampleSearch needs 108 sec for calculating the ground truth for Fig. 13e (The maximum lineage size among all 25 query answers is 1941 in this scenario). Upper dissociation bounds can be calculated in only 3 sec and give the same ranking (and almost the same probabilities).

---

[25]The reported times are for evaluating all answer DNFs without the overhead for the lineage query.

*Key take-aways*: Overall, our quality and timing experiments suggest that dissociation bounds (in particular, the upper bounds) can provide a good approximation of the actual probabilities and provide a ranking of the query answers that is often identical to the ranking for their actual probabilities. These bounds can be calculated with guaranteed polynomial scalability in the size of the data. In particular for queries with small lineage sizes ($< 10000$), calculating the bounds took only a small multiple ($< 4$) of the time needed to evaluate standard deterministic queries.　■

## 8. RELATED WORK AND DISCUSSION

Dissociation is related to a number of recent approaches in the graphical models and constraint satisfaction literature which approximate an intractable problem with a tractable relaxed version after *treating multiple occurrences of variables or nodes as independent* or ignoring some equivalence constraints: Choi et al. [2007] approximate inference in Bayesian networks by "node splitting," i.e. removing some dependencies from the original model. Ramírez and Geffner [2007] treat the problem of obtaining a minimum cost satisfying assignment of a CNF formula by "variable renaming," i.e. replacing a variable that appears in many clauses by many fresh new variables that appear in few. Pipatsrisawat and Darwiche [2007] provide lower bounds for MaxSAT by "variable splitting," i.e. compiling a relaxation of the original CNF. Andersen et al. [2007] improve the relaxation for constraint satisfaction problems by "refinement through node splitting," i.e. making explicit some interactions between variables. Choi and Darwiche [2009] relax the maximum a posteriori (MAP) problem in probabilistic graphical models by dropping equivalence constraints and partially compensating for the relaxation. Our work provides a general framework for approximating the probability of Boolean functions with both upper and lower bounds. These bounds shed light on the connection between previous relaxation-based and model-based approximations and unify them as concrete choices in a larger design space. We thus refer to all of the above approaches as instances of *dissociation-based approximations*.

Another line of work that is varyingly called "discretization," "bucketing," "binning," or "quantization" proposes relaxations by *merging or partitioning states or nodes* (instead of splitting them), and to then perform simplified calculations over those partitions: Dechter and Rish [2003] approximate a function with high arity by a collection of smaller-arity functions with a parameterized scheme called "mini-buckets." As example, the sum $\sum_i f(x_i) \cdot g(x_i)$ for non-negative functions $f$ and $g$ can be upper bounded by the summation $(\max_i f(x_i)) \cdot \sum_i g(x_i)$, i.e. all different values of $f(x_i)$ are replaced by the single maximum value $\max_i f(x_i)$, which simplifies the calculations. Similarly, the sum can be lower bounded by $(\min_i f(x_i)) \cdot \sum_i g(x_i)$. St-Aubin et al. [2000] use Algebraic Decision Diagrams (ADDs) and reduce the sizes of the intermediate value functions generated by replacing the values at the terminals of the ADD with ranges of values. Bergman et al. [2011],[2013] construct relaxations of Multivalued Decision Diagrams (MDDs) by merging vertices when the size of the partially constructed MDD grows too large. Gogate and Domingos [2011] compress potentials computed during the execution of variable elimination by "quantizing" them, i.e. replacing a number of distinct values in range of the potential by a single value. Since all of the above approaches reduce the number of distinct values in the range of a function, we collectively refer to them as *quantization-based approximations*. A more detailed literature overview in this space is given by Gogate and Domingos [2013].

Note that dissociation-based approaches (that split nodes) and quantization-based approaches (that merge nodes) are not inverse operations, but are rather two complementary approaches that may be combined to yield improved methods. The inverse of dissociation is what we refer to as *assimilation*:[26] Consider the Boolean formula $\varphi = x_1 y_1 \vee x_1 y_2 \vee x_2 y_2$ which is a dissociation of the much simpler formula $\varphi^* = x(y_1 \vee y_2)$. Hence, we know from dissociation that $\mathbb{P}[\varphi^*] \leq \mathbb{P}[\varphi]$ for $p \leq \min(p_1, p_2)$ and that $\mathbb{P}[\varphi^*] \geq \mathbb{P}[\varphi]$ for $p \geq 1 - \bar{p}_1 \bar{p}_2$. Note that for quantization, we can always choose $\max(p_1, p_2)$ or $\min(p_1, p_2)$ to get a guaranteed upper or lower bound. In contrast, for assimilation, we may have to choose a different value to get a guaranteed bound. Also, for the case $p_1 = p_2$, assimilation will generally still be an approximation, whereas quantization would be exact. Thus, these are two different approaches.

Existing approaches for query processing over probabilistic databases that are both general and tractable are either: (1) *simulation*-based approaches that adapt general purpose sampling methods [Jampani et al. 2008; Kennedy and Koch 2010; Re et al. 2007]; or (2) *model*-based approaches that approximate the original number of models with guaranteed lower or upper bounds [Olteanu et al. 2010; Fink and Olteanu 2011]. We have show in this paper that, for every model-based bound, there exists a dissociation bound that is at least as good or better.

Our work on dissociation originated while generalizing *propagation-based ranking methods* from graphs [Detwiler et al. 2009] to hypergraphs and conjunctive queries. In [Gatterbauer et al. 2010], we applied dissociation in a query-centric way to upper bound hard probabilistic queries, and showed the connection between propagation on graphs and dissociation on hypergraphs (see [Gatterbauer and Suciu 2013] for all details). In this paper, we provide the theoretical underpinnings of these results in a generalized framework with both upper *and lower bounds*. A previous version of this paper was made available as [Gatterbauer and Suciu 2011].

## 9. OUTLOOK

We introduced dissociation as a new algebraic technique for approximating the probability of Boolean functions. We applied this technique to derive obliviously optimal upper and lower bounds for conjunctive and disjunctive dissociations and proved that dissociation always gives equally good or better approximations than models. We did not address the algorithmic complexities of exploring the space of alternative dissociations, but rather see our technique as a basic building block for new algorithmic approaches. Such future approaches can apply dissociation at two conceptual levels: (1) at the query-level, i.e. at query time and before analyzing the data, or (2) at the data-level, i.e. while analyzing the data.

The advantage of *query-centric* approaches is that they can run in guaranteed polynomial time,[27] yet at the cost of no general approximation guarantees.[28] Here, we envision that the query-centric, first-order logic-based view of operating on data by the database community can also influence neighboring communities, in particular those working on *lifted inference* (see e.g. [Van den Broeck et al. 2011]).

---

[26]We prefer the word "assimilation" as inverse of dissociation over of the more natural choice of "association" as it implies correctly that two items are actually merged and not merely associated.

[27]Recall from our experiments (Fig. 14c) that query-centric dissociation scales linearly in the size of the lineage, independent of intricacies in the data, such as the treewidth of the lineage.

[28]However, also recall from our experiments (Fig. 13) that query-centric approaches may work well *in practice*. Notice here the similarity to loopy belief propagation [Frey and MacKay 1997], which is applied widely and successfully, despite lacking general performance guarantees.

The advantage of *data-centric approaches* is that exact solutions can be arbitrarily approximated, yet at the cost of no guaranteed runtime [Roth 1996]. Here, we envision a range of new approaches (that may combine dissociation with quantization) to compile an existing intractable formula into a tractable target language, e.g., read-once formulas or formulas with bounded treewidth. For example, one can imagine an approximation scheme that adds repeated dissociation to Shannon expansion in order to avoid the intermediate state explosion.

## ACKNOWLEDGMENTS

We would like to thank Arthur Choi and Adnan Darwiche for helpful discussions on Relaxation & Compensation, and for bringing Prop. 5.1 to our attention [Choi and Darwiche 2011]. We would also like to thank Vibhav Gogate for helpful discussions on quantization and guidance for using his tool SampleSearch, Alexandra Meliou for suggesting the name "dissociation," and the reviewers for their careful reading of this manuscript and their detailed feedback. This work was partially supported by NSF grants IIS-0915054 and IIS-1115188. More information about this research, including the PostgreSQL statements to repeat the experiments on TPC-H data, can be found on the project page: http://LaPushDB.com/.

## A. NOMENCLATURE

| | |
|---|---|
| $x, y, z$ | independent Boolean random variables |
| $\varphi, \psi$ | Boolean formulas, probabilistic event expressions |
| $f, g, f_\varphi$ | Boolean function, represented by an expression $\varphi$ |
| $\mathbb{P}[x], \mathbb{P}[\varphi]$ | probability of an event or expression |
| $p_i, q_j, r_k$ | probabilities $p_i = \mathbb{P}[x_i]$, $q_j = \mathbb{P}[y_j]$, $r_k = \mathbb{P}[z_k]$ |
| $\mathbf{x}, \mathbf{g}, \mathbf{p}$ | sets $\{x_1, \ldots, x_k\}$ or vectors $\langle x_1, \ldots, x_k \rangle$ of variables, functions or probabilities |
| $\mathbb{P}_{\mathbf{p}, \mathbf{q}}[f]$ | probability of function $f(\mathbf{x}, \mathbf{y})$ for $\mathbf{p} = \mathbb{P}[\mathbf{x}]$, $\mathbf{q} = \mathbb{P}[\mathbf{y}]$ |
| $\bar{x}, \bar{\varphi}, \bar{p}$ | complements $\neg x, \neg \varphi, 1 - p$ |
| $f', \varphi'$ | dissociation of a function $f$ or expression $\varphi$ |
| $\theta$ | substitution $\theta : \mathbf{x}' \to \mathbf{x}$; defines a dissociation $f'$ of $f$ if $f'[\theta] = f$ |
| $f[x'/x]$ | substitution of $x'$ for $x$ in $f$ |
| $m, m', n$ | $m = |\mathbf{x}|$, $m' = |\mathbf{x}'|$, $n = |\mathbf{y}|$ |
| $d_i$ | number of new variables that $x_i$ is dissociated into |
| $\nu$ | valuation or truth assignment $\nu : \mathbf{y} \to \{0, 1\}$ with $y_i = \nu_i$ |
| $f[\nu], \varphi[\nu]$ | function $f$ or expression $\varphi$ with valuation $\nu$ substituted for $\mathbf{y}$ |
| $\mathbf{g}^\nu$ | $\mathbf{g}^\nu = \bigwedge_j g_j^\nu$, where $g_j^\nu = \bar{g}_j$ if $\nu_j = 0$ and $g_j^\nu = g_j$ if $\nu_j = 1$ |

## B. REPRESENTING COMPLEX EVENTS (DISCUSSION OF COROLLARY 4.5)

It is known from Poole's independent choice logic [Poole 1993] that arbitrary correlations between events can be composed from *disjoint-independent events* only. A disjoint-independent event is represented by a non-Boolean independent random variable $y$ which takes either of $k$ values $\mathbf{v} = \langle v_1, \ldots, v_k \rangle$ with respective probabilities $\mathbf{q} = \langle q_1, \ldots, q_k \rangle$ and $\sum_i q_i = 1$. Poole writes such a "disjoint declaration" as $y([v_1 : q_1, \ldots, v_k : q_k])$.

In turn, any $k$ disjoint events can be represented starting from $k - 1$ independent Boolean variables $\mathbf{z} = \langle z_1, \ldots, z_{k-1} \rangle$ and probabilities $\mathbb{P}[\mathbf{z}] = \langle q_1, \frac{q_2}{\bar{q}_1}, \frac{q_3}{\bar{q}_1 \bar{q}_2}, \ldots, \frac{q_{k-1}}{\bar{q}_1 \ldots \bar{q}_{k-2}} \rangle$, by assigning the disjoint-independent event variable $y$ its value $v_i$ whenever event $A_i$ is true, with $A_i$ defined as:

$$(y = v_1) \equiv A_1 :- z_1$$
$$(y = v_2) \equiv A_2 :- \bar{z}_1 z_2$$
$$\vdots$$
$$(y = v_{k-1}) \equiv A_{k_1} :- z_1 \ldots \bar{z}_{k-2} z_{k-1}$$
$$(y = v_k) \equiv A_k :- \bar{z}_1 \ldots \bar{z}_{k-2} \bar{z}_{k-1} .$$

For example, a disjoint-independent event $y(v_1 : \frac{1}{5}, v_2 : \frac{1}{2}, v_3 : \frac{1}{5}, v_4 : \frac{1}{10})$ can be represented with three independent Boolean variables $\mathbf{z} = (z_1, z_2, z_3)$ and $\mathbb{P}[\mathbf{z}] = (\frac{1}{5}, \frac{5}{8}, \frac{2}{3})$.

It follows that *arbitrary correlations between events* can be modeled starting from *independent Boolean random variables* alone. For example, two *complex events* $A$ and $B$ with $\mathbb{P}[A] = \mathbb{P}[B] = q$ and varying correlation (see Sect. 7.1) can be represented as *composed events* $A :- z_1 z_2 \vee z_3 \vee z_4$ and $B :- \bar{z}_1 z_2 \vee z_3 \vee z_5$ over the *primitive events* $\mathbf{z}$ with varying probabilities $\mathbb{P}[\mathbf{z}]$. Events $A$ and $B$ become identical for $\mathbb{P}[\mathbf{z}] = (0, 0, q, 0, 0)$, independent for $\mathbb{P}[\mathbf{z}] = (0, 0, 0, q, q)$, and disjoint for $\mathbb{P}[\mathbf{z}] = (0.5, q, 0, 0, 0)$ with $q \leq 0.5$.

## C. USER-DEFINED AGGREGATE IOR (SEC. 6 AND SEC. 7.5)

Here we show the User-defined Aggregate (UDA) IOR in PostgreSQL:

```
create or replace function ior_sfunc(float, float) returns float as
  'select $1 * (1.0 - $2)'
  language SQL;

create or replace function ior_finalfunc(float) returns float as
  'select 1.0 - $1'
  language SQL;

create aggregate ior (float)(
  sfunc = ior_sfunc,
  stype = float,
  finalfunc = ior_finalfunc,
  initcond = '1.0');
```